\documentclass[12pt]{article}

\usepackage[T1]{fontenc}
\usepackage[margin=0.8in]{geometry}
\usepackage{microtype}
\usepackage{amsmath,amssymb,bm,mathtools}
\usepackage{booktabs}
\usepackage{longtable}
\usepackage{graphicx}
\usepackage{float}
\usepackage{placeins}
\usepackage{xcolor}
\usepackage{hyperref}
\usepackage{enumitem}

\hypersetup{
  colorlinks=true,
  linkcolor=blue!55!black,
  citecolor=blue!55!black,
  urlcolor=blue!55!black
}

\newcommand{\R}{\mathbb{R}}

\newcommand{\uvec}{\bm{u}}

\begin{document}

\begin{center}
{\Large A differentiable Lagrangian-coupled 3D Gaussian Splatting--SPH model for forward simulation and inverse analysis in solid mechanics\par}

\vspace{1.2em}

{\large Tian Xu, \qquad Soroush Atashi, \qquad Tianju Xue$^{*}$\footnotemark[1]\par}

\vspace{0.8em}

{\itshape
Department of Civil and Environmental Engineering, Hong Kong University of Science and Technology
\par}

\footnotetext[1]{Corresponding author: Tianju Xue. Email: cetxue@ust.hk}

\end{center}

\vspace{1.2em}
\hrule
\vspace{1em}

\noindent{\Large\bfseries Abstract}

\vspace{0.6em}

\noindent 
Recent advances in generative world models for physical artificial intelligence have increased interest in digital models that reproduce both the appearance of real objects and their response to physical interaction.  Three-dimensional reconstruction techniques, including 3D Gaussian Splatting, capture detailed surface geometry and appearance from images and videos, while physics-integrated representations extend these capabilities to simulating the mechanical response of deformable solids.
However, extending these representations beyond plausible animation to mechanically interpretable models for constitutive behavior, boundary conditions, and inverse parameter identification remains less explored.
In this work, a differentiable Lagrangian-coupled 3D Gaussian Splatting (3DGS)--smoothed particle hydrodynamics (SPH) model is proposed for forward simulation and inverse analysis of deformable solids.  The observed object is first reconstructed from multi-view calibrated visual dataset as a 3DGS rendering model.  An envelope-based procedure then generates an independent SPH support for the solid-mechanics model, avoiding the direct use of rendering primitives as mechanical particles. A reference-configuration Lagrangian transfer maps SPH deformation to Gaussian positions and covariances, thereby coupling the physical model and the image observation model while preserving a differentiable computational path.  The SPH formulation supports linear elastic, hyperelastic, and Kelvin--Voigt viscoelastic responses, together with fixed, free, and Robin-type boundary conditions.  Numerical studies validate the SPH response against finite-element results, assess accuracy and efficiency against a conventional model using Gaussian centers as surface SPH particles, and demonstrate forward simulations on beam, bridge, and liver-shaped examples.  Inverse analyses further estimate constitutive and boundary parameters from rendered deformation observations, including noisy cases, demonstrating the feasibility of the proposed model for mechanics-based parameter identification from image data.
\vspace{0.8em}

\vspace{0.8em}

\vspace{0.8em}
\hrule
\section{Introduction}

Recent advances in generative world models for physical artificial
intelligence are extending visual scene generation toward interactive
simulation for robot learning and action evaluation
\cite{nvidia2025worldsimulation}.  Alongside these developments,
reconstructing deformable objects from videos for subsequent physical
simulation is bringing vision and mechanics into closer connection
\cite{jiang2025phystwin}.  Translating these capabilities into mechanical analysis requires more than visually plausible motion: the deformation must be governed by constitutive relations and balance laws under specified boundary conditions, with physical parameters that can be inferred from observations \cite{willcox2021imperative}.  This requires a physical model of the solid coupled with a rendering model that relates its deformation to the observed appearance.  In forward simulation, these models describe the mechanical response and its rendered appearance; in inverse analysis, they enable physical parameters to be estimated by comparing simulated and observed
images.  A consistent and differentiable connection between vision and solid mechanics is therefore needed to support both deformation simulation and parameter identification from images or videos.

Modern visual reconstruction methods provide the geometric and appearance
description required to start from images.  In particular, 3D Gaussian Splatting
(3DGS) represents a scene by anisotropic Gaussian primitives and has become an
effective representation for view synthesis \cite{kerbl2023gaussian}.  A 3DGS
model, however, is primarily an observation-space representation.  Its Gaussian
centers, covariances, colors, and opacities are optimized for rendering, and do
not by themselves define a constitutive model, material volume, or boundary
condition for solid-mechanics analysis.

Physics-integrated Gaussian models have made important progress in forward
deformation simulation of 3DGS representations.  PhysGaussian couples
co-located Gaussians to the material point method (MPM), while Spring-Gaus uses
a spring--mass representation; subsequent methods broadened the constitutive
range to learned or mixed material responses
\cite{xie2024physgaussian,zhong2024springgaus,cao2024neuma,
lin2025omniphysgs,huang2026gaussianfluent}.  These formulations provide a direct kinematic link between simulation and rendering and can produce plausible deformation under prescribed material and loading settings.  
For solid mechanics, however, co-locating mechanical supports with visual primitives would result in a high-dimensional physical discretization, with the mechanical supports tending to be distributed non-uniformly and densely over the object surface.

Several recent methods therefore use reduced or proxy physical representations to improve the efficiency of forward deformation.  Sparse controls can drive dense Gaussians without a constitutive model \cite{huang2024scgs}; PhysSplat employs adaptively sampled MPM driving particles, with material property distributions supplied by multimodal language model priors; PHYSPLAT derives appearance agnostic simulation proxies from reconstructed Gaussian scenes; and GASP groups or selects Gaussians for an external physics engine
\cite{zhao2025physsplat,alfonso2026physplat,borycki2026gasp}.  Most closely, PhysTalk fixes rest pose particle neighborhoods and blends particle displacements and deformation gradients back to the original Gaussians \cite{collorone2025phystalk}.  These studies show that reduced physical representations are promising for efficient forward animation.  
Nevertheless, the physical representation in the aforementioned methods often serves as a proxy or control model for rendered motion. Without adequate constitutive and boundary modeling, visual agreement alone cannot establish mechanical accuracy under different loading conditions.

Beyond computational efficiency and physical fidelity in forward simulation, a further objective is to identify the constitutive parameters and boundary conditions responsible for the observed deformation. 
Full-field methods such as finite-element model updating and the
virtual fields method have established the value of spatially resolved
observations for constitutive identification
\cite{avril2008fullfield}.  Related elasticity inverse problem studies have
shown that material parameters and boundary conditions can be jointly identified
from partial or local observations when the forward mechanics model and
observation operator are properly formulated
\cite{xu2022fembased,xu2023multiple}.  These methods commonly use displacement
or strain fields extracted before the inverse solve.  In rendered inverse
physics, PAC-NeRF couples neural rendering with differentiable MPM for system
identification from multiview videos; PhysDreamer and NeuMA use image-space
supervision to infer material behavior; and TopoGaussian, PhysTwin, and
MonoPhysics estimate topology, physical digital-twin parameters, or joint
geometry--appearance--physics states from visual observations
\cite{li2023pacnerf,zhang2024physdreamer,cao2024neuma,
xiong2025topogaussian,jiang2025phystwin,rho2026monophysics}.  These works
confirm the relevance of rendered identification.  However, examples such as
TopoGaussian mainly emphasize internal topology or material related recovery,
whereas boundary condition identification and its coupling with constitutive
parameters remain less developed in mechanics-oriented Gaussian deformation
models.

For a differentiable model supporting both forward simulation and inverse analysis, the solid discretization and the evaluation of parameter sensitivities must be considered together.
Smoothed Particle Hydrodynamics (SPH) provides a suitable physical
discretization for this purpose because it is mesh-free and naturally
Lagrangian compared to the MPM \cite{lucy1977sph,gingold1977sph,monaghan1992sph}.  Corrected and
corotated SPH formulations further improve the consistency and stability of
elastic solid simulation
\cite{randles1996sph,bonet1999variational,becker2009corotated,peer2018implicit}.
Extending these solid discretization formulations to gradient-based inverse analysis further requires sensitivities of the simulated response to the unknown physical parameters.  Differentiable numerical frameworks provide this capability: JAX-FEM supports sensitivity analysis and inverse design in solid mechanics
\cite{xue2023jaxfem}, while JAX-SPH enables differentiation through SPH fluid simulations for inverse analysis \cite{toshev2024jaxsph}.  However, differentiable SPH formulations for solid mechanics remain less explored, particularly for the joint identification of constitutive parameters and boundary conditions.

Despite these advances, constructing mechanical models from visual observations remains challenging when efficient forward simulation must preserve physical and rendering fidelity with fewer mechanical degrees of freedom. Extending the same model to inverse analysis poses the additional challenge of identifying constitutive parameters and boundary conditions from images within a unified differentiable formulation.
To address these challenges, this work develops a
Lagrangian-coupled 3D Gaussian Splatting--SPH model for forward simulation and
inverse analysis in solid mechanics.  In the proposed model, the observed object is first
represented by a 3D Gaussian Splatting (3DGS) model reconstructed from
multi-view images or video frames.  An envelope-based procedure is then used
to generate an SPH particle support from the reconstructed Gaussian geometry,
while the solid-mechanics response is computed on the SPH model rather than
directly on the Gaussian primitives.  The deformation of the SPH model is
transferred back to the Gaussian representation through a
reference-configuration Lagrangian mapping, which reconstructs Gaussian
centers and covariance deformation from neighboring SPH particle displacement
and local deformation.  The coupled simulation, transfer, and rendering
process is implemented in JAX, allowing the same model to be used for both
forward deformation rendering and gradient-based inverse analysis.

The main contributions of this work are reflected in three aspects.  First, a differentiable Lagrangian SPH-GS transfer is proposed to bridge the 3DGS rendering model and the SPH physical model.  This enables a lower-dimensional mechanical discretization to drive a dense Gaussian representation, improving simulation efficiency while retaining physical and rendering fidelity. Second, the corrected corotated SPH model is extended to incorporate hyperelastic and Kelvin--Voigt viscoelastic responses together with Robin-like
boundary conditions, expanding the physical modeling capabilities of the coupled model beyond elastic deformation under fixed or free constraints. Third, forward simulation and inverse analysis are connected within the same differentiable 3DGS-SPH model, enabling joint identification of constitutive parameters and boundary conditions from image or video observations. The JAX implementation provides parameter gradients through simulation, Lagrangian transfer, and rendering to support gradient-based optimization.

The remainder of this paper is organized as follows.  Section~2 presents the
mathematical formulation of the forward simulation and inverse analysis
problems.  Section~3 introduces the proposed 3DGS-SPH model, including the
3D Gaussian representation, envelope-based SPH support generation, corrected
solid SPH formulation, Lagrangian SPH-GS transfer, and inverse
analysis formulation.  Section~4 reports numerical experiments for SPH
validation, mapping assessment, forward simulation, sensitivity verification,
and inverse parameter identification.  Section~5 discusses the numerical
results, and Section~6 concludes the paper with
potential directions for future work.


\section{Problem Statement}
\label{sec:problem-formulation}
\begin{figure}[htbp!]
  \centering
\includegraphics[width=0.8\textwidth]{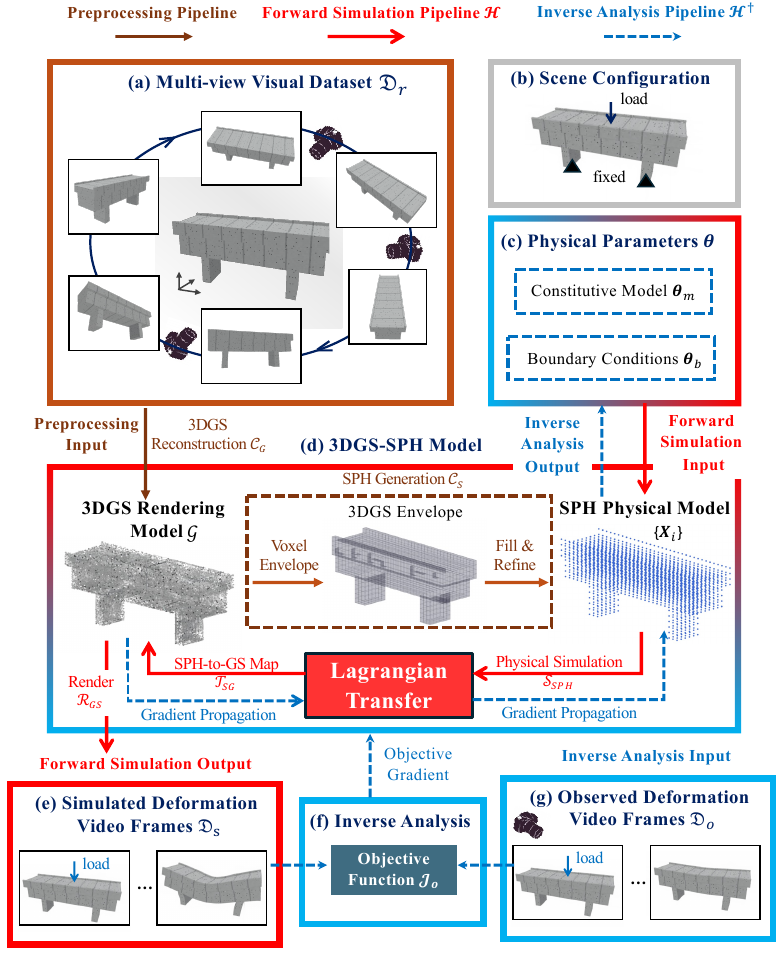}
  \caption{Workflow of the proposed Lagrangian-coupled 3DGS-SPH model for
  forward simulation and inverse analysis.  (a) Reference multiview images;
  (b) prescribed scene configuration; (c) constitutive and boundary parameters;
  (d) 3DGS rendering model and SPH physical model coupled through the proposed
  Lagrangian transfer; (e) simulated deformation frames; (f) inverse analysis
  using the discrepancy between simulated and observed frames; and
  (g) observed deformation frames.  Brown arrows indicate model construction,
  red arrows indicate forward simulation, and dashed blue arrows indicate
  inverse analysis, including gradient propagation through the coupled model
  for parameter estimation.}
  \label{fig:framework-workflow}
\end{figure}

Consider an opaque solid object observed in a reference configuration and during deformation.  The observations comprise the reference dataset \(\mathfrak{D}_{\mathrm{r}}=\{\mathcal{Y}_m^{\mathrm{r}}\}_m\), acquired from multiple calibrated views in the reference configuration (Fig.~\ref{fig:framework-workflow}(a)), and the deformation dataset \(\mathfrak{D}_{\mathrm{o}}=\{\mathcal{Y}^{\mathrm{o}}(t)\}_t\), recorded from a single fixed calibrated view during deformation (Fig.~\ref{fig:framework-workflow}(g)).  Here
\(\mathcal{Y}_m^{\mathrm{r}}\) denotes the reference image from view \(m\), and \(\mathcal{Y}^{\mathrm{o}}(t)\) denotes the deformation image at observation time \(t\).  The physical parameters governing the deformation (Fig.~\ref{fig:framework-workflow}(c)) are collected in \(\bm{\theta}=(\bm{\theta}_m,
\bm{\theta}_b)\in\Theta\), where \(\bm{\theta}_m\) and \(\bm{\theta}_b\) contain constitutive and boundary parameters, respectively, and \(\Theta\) is their admissible parameter domain.  For each analysis, the camera calibration, reference geometry, loading and boundary region definition are prescribed as illustrated in Fig.~\ref{fig:framework-workflow}(b).

Given the reference observations and prescribed scene configuration, the forward
problem is to simulate the observable deformation for specified physical
parameters \(\bm{\theta}\).  Denoting the forward observation operator by
\(\mathcal{H}\), the forward simulation is written as
\begin{equation}
  \mathfrak{D}_{\mathrm{s}}(t;\bm{\theta})
  =\mathcal{H}(\bm{\theta},t;\mathfrak{D}_{\mathrm{r}}),
  \qquad \bm{\theta}\in\Theta,
  \label{eq:problem-forward-map}
\end{equation}
where \(\mathfrak{D}_{\mathrm{s}}(t;\bm{\theta})\) contains the
simulated image at the observation view and time \(t\).  The operator
\(\mathcal{H}\) relates the physical deformation under the prescribed
conditions to these observations; its outputs over the observation times
form the deformation image or video frames.

Conversely, inverse analysis estimates the unknown physical parameters
from \(\mathfrak{D}_{\mathrm{o}}\), with the reference observations and
prescribed conditions retained.  Denoting the inverse estimation operator by
\(\mathcal{H}^{\dagger}\), the parameter estimate is defined by
\begin{equation}
  \begin{aligned}
  \widehat{\bm{\theta}}
  &=\mathcal{H}^{\dagger}
  (\mathfrak{D}_{\mathrm{o}};\mathfrak{D}_{\mathrm{r}})
  \\
  &=
  \underset{\bm{\theta}\in\Theta}{\operatorname{arg\,min}}
  \left\{
    \mathcal{J}_{\mathrm{o}}
    \!\left(
      \{\mathfrak{D}_{\mathrm{s}}(t;\bm{\theta})\}_t,
      \mathfrak{D}_{\mathrm{o}}
    \right)
    +
    \mathcal{R}(\bm{\theta})
  \right\},
  \end{aligned}
  \label{eq:problem-inverse-map}
\end{equation}
where \(\widehat{\bm{\theta}}\) denotes the estimated parameters,
\(\mathcal{J}_{\mathrm{o}}\) measures the discrepancy between simulated
and observed deformation data and \(\mathcal{R}\) represents prior or spatial
regularization.  The simulated images in this objective are supplied by
\(\mathcal{H}\) in Eq.~\eqref{eq:problem-forward-map}.

\section{Method}
\label{sec:method}
To address the forward simulation and inverse analysis problems formulated in Section~\ref{sec:problem-formulation}, this work proposes a Lagrangian-coupled 3DGS-SPH model, as illustrated in
Fig.~\ref{fig:framework-workflow}(d). The proposed model connects a 3DGS rendering model with a separately discretized SPH physical model through a differentiable Lagrangian SPH-GS transfer.

The pipeline begins with reconstruction of the 3DGS
rendering model from the reference images in Fig.~\ref{fig:framework-workflow}(a), represented by the operator \(\mathcal{C}_{\mathrm{G}}\). Since solid mechanics requires a
volumetric discretization, the envelope-based construction operator
\(\mathcal{C}_{\mathrm{S}}\) generates the reference SPH supports by
constructing a voxel envelope from the reconstructed Gaussian geometry,
followed by filling and refinement (Fig.~\ref{fig:framework-workflow}(d)):
\begin{equation}
  \mathcal{G}^0=\mathcal{C}_{\mathrm{G}}(\mathfrak{D}_{\mathrm{r}}),
  \qquad
  \{\bm{X}_i\}_{i=1}^{N}=\mathcal{C}_{\mathrm{S}}(\mathcal{G}^0).
  \label{eq:method-model-construction}
\end{equation}
Here \(\mathcal{G}^0\) contains the reference Gaussian parameters, and
\(\bm{X}_i\) is the reference position of SPH support \(i\) among
\(N\) supports.

The reconstructed 3DGS rendering model and the generated SPH physical model are then coupled through the proposed Lagrangian SPH-GS transfer, giving the forward operator in Eq.~\eqref{eq:problem-forward-map} as
\begin{equation}
  \mathcal{H}(\bm{\theta},t;\mathfrak{D}_{\mathrm{r}})
  =\mathcal{R}_{\mathrm{GS}}
  \!\left[
    \mathcal{T}_{\mathrm{SG}}
    \!\left(
      \mathcal{S}_{\mathrm{SPH}}
      (\bm{\theta},t;\{\bm{X}_i\}_{i=1}^{N});\mathcal{G}^0
    \right)
  \right].
  \label{eq:method-coupled-forward-map}
\end{equation}
The physical evolution operator \(\mathcal{S}_{\mathrm{SPH}}\) determines
the current SPH positions using the scene configuration in
Fig.~\ref{fig:framework-workflow}(b) and physical parameters in
Fig.~\ref{fig:framework-workflow}(c).  The proposed Lagrangian SPH--GS
transfer \(\mathcal{T}_{\mathrm{SG}}\) takes these SPH positions and the
reference Gaussian parameters as input and determines the deformed Gaussian
centers and covariances, thereby connecting the physical and rendering models
in Fig.~\ref{fig:framework-workflow}(d).  The rendering operator
\(\mathcal{R}_{\mathrm{GS}}\) then converts the deformed Gaussian primitives into simulated images at the
observation view (Fig.~\ref{fig:framework-workflow}(e)).

For inverse analysis, these simulated images
are compared with the observed sequence in
Fig.~\ref{fig:framework-workflow}(g) through the objective in
Eq.~\eqref{eq:problem-inverse-map}.  Differentiation through the same coupled
model provides the parameter gradients used to evaluate
\(\mathcal{H}^{\dagger}\) through the inverse solver in
Fig.~\ref{fig:framework-workflow}(f) to obtain the estimates of physical parameters in Fig.~\ref{fig:framework-workflow}(c).

The methodological details are introduced in the following four subsections.
Section~\ref{sec:method-gaussian} describes the 3DGS representation, rendering
operation, and envelope-based construction of the SPH support.
Section~\ref{sec:method-sph-discretization} summarizes the corrected solid-SPH
model, constitutive responses, boundary conditions, and matrix-free solution.
Section~\ref{sec:method-lagrangian-transfer} derives the proposed Lagrangian
SPH-GS transfer, which couples a dense Gaussian representation to an
independently discretized physical model.  Finally,
Section~\ref{sec:method-ad-inverse} formulates parameter identification through
the resulting coupled model.

\subsection{3DGS rendering model}
\label{sec:method-gaussian}

\subsubsection{Reference Gaussian representation and rendering}

The calibrated multi-view observations define a reference 3DGS representation
of the visible object surface.  Following the standard anisotropic formulation
of Kerbl et al.~\cite{kerbl2023gaussian}, let
\(g\in\{1,\ldots,G\}\) index the reconstructed Gaussians, where \(G\) is
their total number.  Reconstruction from the reference dataset
\(\mathfrak{D}_{\mathrm{ref}}\) yields
\begin{equation}
  \mathcal{G}^{0}
  =
  \left\{
    \bm{\mu}_g^0,
    \bm{\Sigma}_g^0,
    o_g,
    \bm{c}_g
  \right\}_{g=1}^{G},
  \qquad
  \bm{\Sigma}_g^0
  =
  \bm{R}_g^0
  \operatorname{diag}(\bm{s}_g^2)
  (\bm{R}_g^0)^T .
  \label{eq:method-reference-reconstruction}
\end{equation}
For Gaussian \(g\), \(\bm{\mu}_g^0\in\R^3\) is its reference center,
\(\bm{\Sigma}_g^0\in\R^{3\times3}\) is its covariance matrix, \(o_g\) is
its opacity coefficient, and \(\bm{c}_g\) contains its spherical-harmonic
appearance coefficients.  The covariance is factorized by the principal scales
\(\bm{s}_g\in\R_+^3\) and orientation \(\bm{R}_g^0\in SO(3)\), as shown in
Eq.~\eqref{eq:method-reference-reconstruction}, and defines the ellipsoidal
rendering support.  For a prescribed camera, each ellipsoid is projected onto
the image plane as a two-dimensional Gaussian.  In the following rendering
equations, \(\bm{\mu}_g\) and \(\bm{\Sigma}_g\) denote the Gaussian center
and covariance at the rendered state; their time index is omitted for clarity.
Let \(\bm{\pi}:\R^3\rightarrow\R^2\) denote the camera projection,
\(\bm{p}\in\R^2\) an image-plane location, and
\(\bm{J}_g\in\R^{2\times3}\) its Jacobian evaluated at \(\bm{\mu}_g\).
The projected covariance and effective opacity are
\begin{equation}
  \bm{\Sigma}_g^{\mathrm{2D}}
  =
  \bm{J}_g\bm{\Sigma}_g\bm{J}_g^T,
  \qquad
  \alpha_g(\bm{p})
  =
  o_g
  \exp\!\left[
    -\frac{1}{2}
    (\bm{p}-\bm{\pi}(\bm{\mu}_g))^T
    (\bm{\Sigma}_g^{\mathrm{2D}})^{-1}
    (\bm{p}-\bm{\pi}(\bm{\mu}_g))
  \right].
  \label{eq:method-projected-gaussian}
\end{equation}
The
rendering operator \(\mathcal R_{\mathrm{GS}}\) in
Eq.~\eqref{eq:method-coupled-forward-map} maps the Gaussian centers and
covariances to an image.  Its pixel values are
\begin{equation}
  \mathcal{Y}(\bm{p})
  =
  \left[\mathcal R_{\mathrm{GS}}
  \!\left(\{\bm\mu_g,\bm\Sigma_g\}_{g=1}^{G}\right)\right](\bm p)
  =
  \sum_{g=1}^{G}
  \alpha_g(\bm{p})
  \bm{c}_g(\bm{d}_g)
  \prod_{\ell=1}^{g-1}
  \left(1-\alpha_{\ell}(\bm{p})\right).
  \label{eq:method-gaussian-render-operator}
\end{equation}
Here \(\ell\) is a dummy index over Gaussians preceding \(g\), so the product is
their accumulated transmittance.  Evaluating
Eq.~\eqref{eq:method-gaussian-render-operator} over all pixels yields the
simulated image introduced in Section~\ref{sec:problem-formulation}.  Within a
fixed visibility order, the rendered colors retain continuous dependence on
the Gaussian centers, covariances, opacities, and appearance coefficients.

\subsubsection{Envelope-based SPH support generation}
\label{sec:method-envelope-filling}

\begin{figure}[htbp!]
  \centering
  \includegraphics[width=0.92\textwidth]{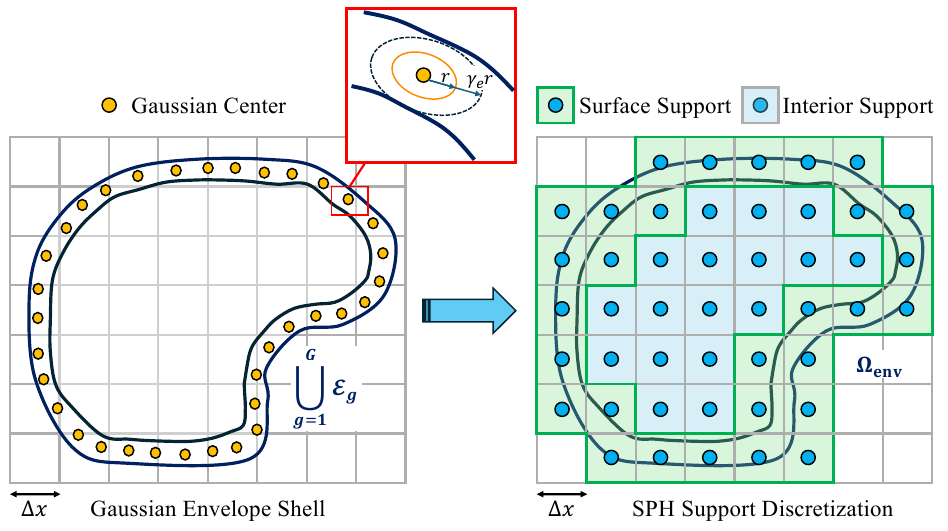}
  \caption{Envelope-based generation of the SPH support from a reconstructed
  Gaussian model.  Expanded Gaussian ellipsoids first define a
  finite-thickness shell.  After gap closing and interior filling, occupied
  grid-cell centers are retained as surface and interior SPH supports.}
  \label{fig:method-envelope-filling}
\end{figure}

Although the Gaussian representation describes the visible surface, solid
mechanics requires a volumetric reference domain.  An envelope-filling
procedure is therefore used to construct the SPH support from the reconstructed
3DGS model.  The insertion of interior supports follows the grid-based
volumetric filling rule used in PhysGaussian and
TopoGaussian~\cite{xie2024physgaussian,xiong2025topogaussian}.  In contrast,
the surface supports are not sampled from the original Gaussian primitives; both
surface and interior supports are defined by occupied grid-cell centers, as
illustrated in Fig.~\ref{fig:method-envelope-filling}.

The construction begins with a confidence ellipsoid \(\mathcal{E}_g\) associated
with each retained Gaussian.  Let \(\gamma_{\mathrm e}>0\) denote the
dimensionless extent factor that scales the Gaussian semi-axes.  The ellipsoid
is defined as
\begin{equation}
  \mathcal{E}_g
  =
  \left\{
    \bm{x}\in\R^3:
    (\bm{x}-\bm{\mu}_g^0)^T
    (\bm{\Sigma}_g^0)^{-1}
    (\bm{x}-\bm{\mu}_g^0)
    \leq \gamma_{\mathrm e}^{2}
  \right\}.
  \label{eq:method-envelope-ellipsoid}
\end{equation}
Because the reconstructed Gaussians are concentrated near the visible surface,
the union of these ellipsoids generally forms a finite-thickness shell rather
than a solid domain.  The occupied reference domain is obtained by closing gaps
in this shell and filling the enclosed interior:
\begin{equation}
  \Omega_{\mathrm{env}}
  =
  \operatorname{Fill}
  \left[
    \operatorname{Close}
    \left(
      \bigcup_{g=1}^{G}\mathcal{E}_g
    \right)
  \right].
  \label{eq:method-envelope-domain}
\end{equation}
Equation~\eqref{eq:method-envelope-domain} is evaluated on a Cartesian grid of
spacing \(\Delta x\), where \(\operatorname{Close}\) denotes morphological
closing of the binary occupancy.  The exterior region is identified by a
boundary-seeded flood fill, and occupied cells adjacent to the exterior are
classified as surface supports~\cite{serra1982image,rosenfeld1966sequential}.
The centers of the resulting occupied cells define the \(N\) reference SPH
supports \(\{\bm{X}_i\}_{i=1}^{N}\).  Consequently, the physical resolution and
support distribution are controlled by \(\Delta x\), rather than by the number
or surface concentration of the reconstructed Gaussians.  Unlike the MPM
formulation adopted in PhysGaussian~\cite{xie2024physgaussian}, this Cartesian
grid is used only to generate the SPH support in the reference configuration;
the subsequent solid-mechanics simulation is performed by the mesh-free SPH
discretization.

\subsection{Solid Lagrangian SPH model}
\label{sec:method-sph-discretization}

The physical simulation operator \(\mathcal{S}_{\mathrm{SPH}}\) in
Eq.~\eqref{eq:method-coupled-forward-map} is based on corrected
total-Lagrangian solid SPH following Peer et al.~\cite{peer2018implicit}.
The adopted formulation is further extended to incorporate neo-Hookean hyperelasticity,
Kelvin--Voigt viscosity, and Robin-like boundary restraint within the same
implicit solution.

For a material point \(\bm X\), the motion
\(\bm x=\bm\chi(\bm X,t)\), deformation gradient \(\bm F\), and balance of
linear momentum in the reference configuration are
\begin{equation}
  \bm{x}=\bm{\chi}(\bm{X},t),
  \qquad
  \bm{F}=\nabla_{\!\bm{X}}\bm{\chi},
  \qquad
  \rho_0\ddot{\bm{\chi}}
  =
  \operatorname{Div}_{\!\bm{X}}\bm{P}
  +\rho_0\bm{b},
  \label{eq:method-continuum-solid}
\end{equation}
where \(\rho_0\) is the prescribed constant reference density,
\(\bm P=\widehat{\bm P}(\bm F,\dot{\bm F};\bm\theta_m)\) is the first
Piola--Kirchhoff stress, \(\dot{\bm F}\) is the material time derivative of
\(\bm F\), \(\bm\theta_m\) collects the material parameters, and \(\bm b\) is
body force per unit mass.  The elastic constitutive models are corotated linear elastic or
neo-Hookean hyperelastic, parameterized by Young's modulus \(E\) and Poisson's ratio
\(\nu\).  Kelvin--Voigt viscosity model adds a rate-dependent stress governed by
the shear and bulk viscosity parameters \(\eta_s\) and \(\eta_b\).
The stress relations are given in Appendix~\ref{sec:method-corrected-sph}.

The reference domain is discretized by \(N\) material supports at
\(\bm X_i\), with reference volumes \(\Delta V_i\) and masses
\(m_i=\rho_0\Delta V_i\), separately from the rendering primitives.
For prescribed initial and boundary conditions, the SPH solution operator
returns their current positions:
\begin{equation}
  \left[
    \mathcal S_{\mathrm{SPH}}
    (\bm\theta,t;\{\bm X_j\}_{j=1}^{N})
  \right]_i
  =\bm x_i(t;\bm\theta)
  =\bm\chi(\bm X_i,t;\bm\theta).
  \label{eq:method-sph-discrete-motion}
\end{equation}
Here \([\cdot]_i\) selects support \(i\), and \(\bm\chi\) at the supports
denotes the discretized motion with physical
parameters \(\bm\theta\).

\begin{figure}[htbp!]
  \centering
  \includegraphics[width=0.58\textwidth]{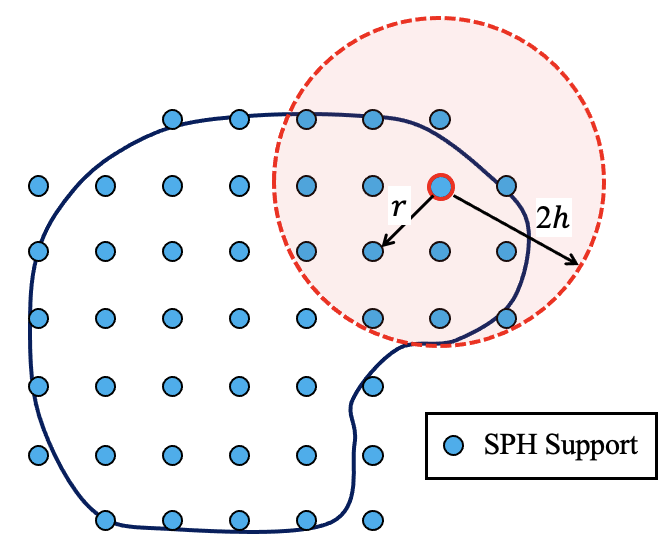}
  \caption{Schematic support relation in the total-Lagrangian SPH
  discretization.  The approximation at a reference support uses neighboring
  supports within the compact kernel radius.}
  \label{fig:method-sph-weight}
\end{figure}

Spatial interactions use a fixed reference graph
\(\mathcal E_{\mathrm{sph}}\), with neighbor set
\(\mathcal N_i=\{j:(i,j)\in\mathcal E_{\mathrm{sph}}\}\).
The neighbor relation is
illustrated in Fig.~\ref{fig:method-sph-weight}.
The same three-dimensional cubic spline kernel is used for the SPH
approximation and the Lagrangian SPH--GS transfer:
\begin{equation}
  W(r,h)
  =
  \frac{1}{\pi h^3}
  \begin{cases}
    1-\frac{3}{2}q^2+\frac{3}{4}q^3, & 0\leq q<1,\\
    \frac{1}{4}(2-q)^3, & 1\leq q<2,\\
    0, & q\geq 2,
  \end{cases}
  \qquad
  q=\frac{r}{h}.
  \label{eq:method-cubic-spline-kernel}
\end{equation}
Here \(r\) is reference distance and \(h\) is the mechanical smoothing
length.  In the SPH model, \(r=\|\bm X_i-\bm X_j\|_2\) relates pairs of
physical supports.  Reference gradient correction provides first-order
consistency, and the corotated formulation removes local rigid rotation
before evaluating linear elastic strain.  The interpolation, gradient
correction, and rotation extraction are detailed in Appendix~\ref{sec:method-corrected-sph}.

The SPH model admits Dirichlet, Neumann, and Robin-like boundary conditions. For prescribed initial conditions, the discrete forward simulation is
expressed as
\begin{equation}
  \mathcal Q(\bm z,\bm\theta)=\bm0,
  \label{eq:method-forward-residual}
\end{equation}
where \(\bm z\) collects the unknown SPH positions and velocities over all
simulated time steps, and \(\bm\theta\) contains the physical parameters.
The residual operator \(\mathcal Q\) assembles the discrete momentum
equations and position updates after imposing the boundary conditions.  The state is
computed sequentially using a matrix-free implicit velocity solver;
the boundary formulation and time integration are detailed in
Appendix \ref{sec:method-time-integration}.  Equation~\eqref{eq:method-forward-residual}
also defines the mechanical constraint used in the inverse gradient
derivation in Section~\ref{sec:method-ad-inverse}.

The resulting SPH positions supply the physical motion for the proposed
Lagrangian SPH--GS transfer, which reconstructs Gaussian motion and local
deformation as described next.

\subsection{Lagrangian SPH-GS transfer}
\label{sec:method-lagrangian-transfer}

The key coupling problem is to deform a dense Gaussian representation using
the motion computed on an independently discretized SPH model.  For each
Gaussian center, the required kinematic quantities are the transported center
and the local deformation gradient of the solid motion:
\begin{equation}
  \bm{\mu}_g(t)
  =
  \bm{\chi}(\bm{\mu}_g^0,t),
  \qquad
  \bm{F}_g(t)
  =
  \nabla_{\!\bm{X}}\bm{\chi}(\bm{\mu}_g^0,t),
  \qquad
  \bm{\Sigma}_g(t)
  =
  \bm{F}_g(t)\bm{\Sigma}_g^0\bm{F}_g(t)^T .
  \label{eq:method-transfer-continuum}
\end{equation}
Here \(\bm{\mu}_g^0\), \(\bm{\mu}_g(t)\), and \(\bm{\Sigma}_g(t)\) are the
reference center, deformed center, and deformed covariance of Gaussian \(g\),
and \(\bm{F}_g(t)\) is the deformation gradient sampled at
\(\bm{\mu}_g^0\).  This relation is also used by co-located
physics-integrated Gaussian models, where Gaussian samples are tied to
mechanical particles and the motion can be obtained directly from the
simulation~\cite{xie2024physgaussian,xiong2025topogaussian}.  Such a
construction, however, can pass the surface-biased density and resolution of
the reconstructed appearance model to the mechanical discretization.  The
proposed model keeps the SPH supports independent from the Gaussian centers and
therefore reconstructs \(\bm{\chi}(\bm{\mu}_g^0,t)\) and
\(\nabla_{\!\bm X}\bm{\chi}(\bm{\mu}_g^0,t)\) from the SPH trajectory using a
fixed reference support graph, normalized transfer weights, and a local affine
fit.

\subsubsection{Lagrangian support graph and transfer weights}

\begin{figure}[htbp!]
  \centering
  \includegraphics[width=0.62\textwidth]{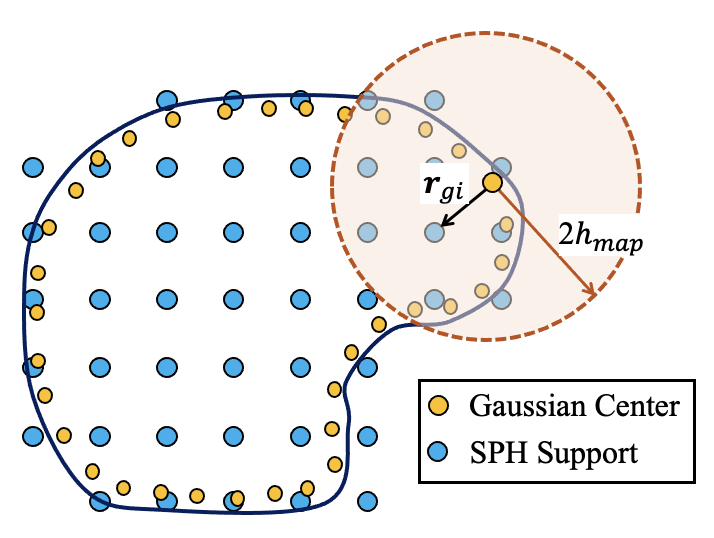}
  \caption{Reference-configuration support relation for the Lagrangian
  SPH-GS transfer.  Each Gaussian center selects neighboring SPH
  supports within the transfer kernel radius, forming a fixed bipartite graph
  for deformation transfer.}
  \label{fig:method-SPH--GS-weight}
\end{figure}

For each reference Gaussian center, a fixed support graph is built in the
reference configuration by selecting neighboring SPH supports within a transfer
radius \(\bar h_{\mathrm{map}}\):
\begin{equation}
  \mathcal{E}_{\mathrm{map}}
  =
  \left\{
    (g,i):
    \|\bm{\mu}_g^0-\bm{X}_i\|_2<\bar h_{\mathrm{map}}
  \right\},
  \qquad
  \mathcal{N}_g
  =
  \{i:(g,i)\in\mathcal{E}_{\mathrm{map}}\}.
  \label{eq:method-map-graph}
\end{equation}
Here \(r_{gi}=\|\bm{\mu}_g^0-\bm{X}_i\|_2\) denotes the reference distance
between Gaussian \(g\) and SPH support \(i\), and \(\mathcal{N}_g\) is the
set of SPH supports used to transfer motion to that Gaussian.  Because the
graph is constructed from reference positions and then kept fixed, the
coupling is Lagrangian rather than an Eulerian re-search during deformation.
Unlike the mechanical graph \(\mathcal{E}_{\mathrm{sph}}\), which connects SPH
support pairs for force evaluation, \(\mathcal{E}_{\mathrm{map}}\) connects
Gaussian centers to SPH supports only for representation transfer.  The
geometric meaning of this support relation is shown in
Fig.~\ref{fig:method-SPH--GS-weight}.

The transfer weights are constructed from the same cubic-spline kernel
\(W(r,h)\) defined in Eq.~\eqref{eq:method-cubic-spline-kernel}.  This choice
gives the Gaussian-to-SPH transfer the same compact and distance-decaying
smoothing profile as the SPH approximation, while changing the relation to
which the kernel is applied: Eq.~\eqref{eq:method-sph-interpolation} relates
one SPH support to neighboring SPH supports, whereas the present transfer
relates one Gaussian center to neighboring SPH supports.  With
\(h_{\mathrm{map}}\) as the transfer smoothing length, the compact support of
Eq.~\eqref{eq:method-cubic-spline-kernel} gives
\(\bar h_{\mathrm{map}}=2h_{\mathrm{map}}\).  The normalized transfer weights
are
\begin{equation}
  \omega_{gi}
  =
  \frac{
    W_{h_{\mathrm{map}}}(\bm{\mu}_g^0-\bm{X}_i)
  }{
    \displaystyle
    \sum_{j\in\mathcal{N}_g}
    W_{h_{\mathrm{map}}}(\bm{\mu}_g^0-\bm{X}_j)
  },
  \qquad
  \sum_{i\in\mathcal{N}_g}\omega_{gi}=1.
  \label{eq:method-map-weights}
\end{equation}
where \(W_{h_{\mathrm{map}}}(\bm r)=W(\|\bm r\|_2,h_{\mathrm{map}})\).  The
normalization gives the transfer a constant-field reproduction property.  


\subsubsection{Transfer of motion and local deformation}

Let \(n\) denote the time-step index, and let
\(\bm u_i^n=\bm x_i^n-\bm X_i\) be the displacement of SPH support \(i\).  Once
the transfer weights have been fixed, the deformed Gaussian center is obtained
by the weighted SPH displacement:
\begin{equation}
  \bm{\mu}_g^n
  =
  \bm{\mu}_g^0
  +
  \sum_{i\in\mathcal{N}_g}
  \omega_{gi}(\bm{x}_i^n-\bm{X}_i).
  \label{eq:method-gaussian-center-transfer}
\end{equation}
Here \(\bm\mu_g^n\) is the center of Gaussian \(g\) at time step \(n\).
Center interpolation alone cannot determine the rotation, stretch, and shear
of an anisotropic Gaussian.  The Gaussian-level deformation gradient
\(\bm F_g^n\in\R^{3\times3}\) is therefore reconstructed from the same support
graph by matching the relative positions of neighboring SPH supports in the
reference and deformed configurations:
\begin{equation}
  \bm{F}_g^n
  =
  \underset{\bm{F}\in\R^{3\times3}}{\operatorname{arg\,min}}
  \left\{
  \sum_{i\in\mathcal{N}_g}
  \omega_{gi}
  \left\|
    (\bm{x}_i^n-\bm{\mu}_g^n)
    -
    \bm{F}(\bm{X}_i-\bm{\mu}_g^0)
  \right\|_2^2
  +
  \lambda_{\mathrm{map}}\|\bm{F}-\bm{I}\|_F^2
  \right\}.
  \label{eq:method-map-affine-fit}
\end{equation}
In Eq.~\eqref{eq:method-map-affine-fit}, \(\bm F\) is the trial affine
deformation tensor, \(\|\cdot\|_2\) is the Euclidean norm, \(\|\cdot\|_F\) is
the Frobenius norm, \(\bm I\) is the identity tensor, and
\(\lambda_{\mathrm{map}}\geq0\) is a small regularization parameter that keeps
the affine reconstruction well-conditioned when the local support geometry is
nearly degenerate.  Defining the reference offset
\(\bm{r}_{gi}=\bm{X}_i-\bm{\mu}_g^0\), the minimizer has the closed form
\begin{equation}
  \begin{aligned}
  \bm{B}_g
  &=
  \sum_{i\in\mathcal{N}_g}
  \omega_{gi}\bm{r}_{gi}\otimes\bm{r}_{gi}
  +\lambda_{\mathrm{map}}\bm{I},
  \\
  \bm{A}_g^n
  &=
  \sum_{i\in\mathcal{N}_g}
  \omega_{gi}
  (\bm{x}_i^n-\bm{\mu}_g^n)
  \otimes\bm{r}_{gi}
  +\lambda_{\mathrm{map}}\bm{I},
  \qquad
  \bm{F}_g^n=\bm{A}_g^n\bm{B}_g^{-1}.
  \end{aligned}
  \label{eq:method-map-F}
\end{equation}
Here \(\bm B_g\) is the regularized reference moment matrix of the transfer
support around Gaussian \(g\), and \(\bm A_g^n\) is the corresponding
deformed cross-moment matrix.  Since \(\bm B_g\) depends only on the reference
configuration and fixed weights, \(\bm B_g^{-1}\) is precomputed.  At the
reference configuration, \(\bm{A}_g^0=\bm{B}_g\) and
\(\bm{F}_g^0=\bm{I}\).  The deformed covariance is then obtained from the
affine push-forward in Eq.~\eqref{eq:method-transfer-continuum} and rendered
by \(\mathcal{R}_{\mathrm{3DGS}}\) in
Eq.~\eqref{eq:method-gaussian-render-operator}.

\subsubsection{Differentiability and relation to co-located coupling}

At time step \(n\), the Lagrangian SPH-GS transfer introduced in
Eq.~\eqref{eq:method-coupled-forward-map} maps the SPH positions to Gaussian
centers and covariances:
\begin{equation}
  \mathcal{T}_{\mathrm{SG}}^n:
  \{\bm{x}_i^n\}_{i=1}^{N}
  \longmapsto
  \left\{
    \bm{\mu}_g^n,
    \bm{\Sigma}_g^n
  \right\}_{g=1}^{G},
  \label{eq:method-forward-transfer-operator}
\end{equation}
This mapping comprises the weighted center transfer in
Eq.~\eqref{eq:method-gaussian-center-transfer}, the local affine
reconstruction of \(\bm F_g^n\) in
Eqs.~\eqref{eq:method-map-affine-fit} and
\eqref{eq:method-map-F}, and the covariance push-forward in
Eq.~\eqref{eq:method-transfer-continuum}.  Its reference quantities
\(\{\bm X_i\}_{i=1}^{N}\), \(\{\bm\mu_g^0,\bm\Sigma_g^0\}_{g=1}^{G}\), graph
\(\mathcal{E}_{\mathrm{map}}\), and weights \(\omega_{gi}\) are fixed;
the resulting transfer is differentiable with respect to the SPH positions.

For inverse analysis, differentiation of the image objective through the
renderer yields sensitivities with respect to the Gaussian centers and
covariances.  These sensitivities are propagated to the SPH positions through
the adjoint of the transfer differential:
\begin{equation}
  \left(
    D_{\bm{x}^n}
    \mathcal{T}_{\mathrm{SG}}^n
  \right)^*:
  \left\{
    \overline{\bm{\mu}}_g^n,
    \overline{\bm{\Sigma}}_g^n
  \right\}_{g=1}^{G}
  \longmapsto
  \left\{
    \overline{\bm{x}}_{i,\mathrm{map}}^n
  \right\}_{i=1}^{N},
  \label{eq:method-map-adjoint-operator}
\end{equation}
where \(D_{\bm x^n}\mathcal T_{\mathrm{SG}}^n\) is the differential with
respect to the SPH positions, \((\cdot)^*\) denotes its adjoint, and an
overbar denotes an objective sensitivity (cotangent).  The subscript
\(\mathrm{map}\) identifies the contribution from the transfer to each
SPH position sensitivity. It is evaluated automatically by reverse-mode
differentiation of the forward transfer which is built based on JAX.  The resulting SPH sensitivities then enter the mechanical adjoint to obtain material and boundary parameter gradients, as detailed in Section~\ref{sec:method-ad-inverse}.

Compared with co-located coupling, where Gaussian-derived samples are tied to
mechanical particles, the proposed transfer permits many Gaussians to share a
smaller set of SPH supports while retaining distinct weights and locally
reconstructed deformation gradients.  It therefore decouples rendering resolution from the dimension and surface concentration of the mechanical system while providing the sensitivity propagation required for inverse analysis.

\subsection{Inverse analysis with the coupled 3DGS-SPH model}
\label{sec:method-ad-inverse}

\begin{figure}[htbp!]
  \centering
  \includegraphics[width=0.8\textwidth]{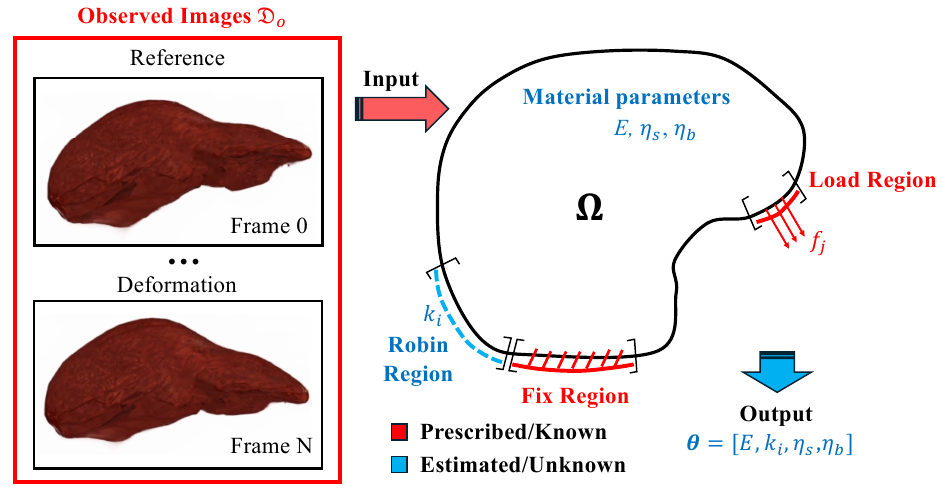}
  \caption{Schematic setting for inverse analysis.  The observed image sequence is
  used to identify selected physical parameters, including elastic modulus,
  Robin boundary stiffness, and, for dynamic viscoelastic cases,
  Kelvin--Voigt viscosity parameters.}
  \label{fig:method-inverse-problem}
\end{figure}

The inverse estimation operator \(\mathcal{H}^{\dagger}\) is evaluated
through the minimization in Eq.~\eqref{eq:problem-inverse-map} to estimate
selected physical parameters from deformation images.  The \(\bm{\theta}\) denote the unknown parameter vector, which contain the
Young's modulus \(E\), Robin stiffness values \(k_i\),  Kelvin--Voigt parameters \(\eta_s\) and \(\eta_b\), as
summarized in Fig.~\ref{fig:method-inverse-problem}.  Positive physical
quantities are represented by dimensionless logarithmic variables,
\begin{equation}
  p=p_{\mathrm{r}}\exp(\theta_p),
  \qquad
  p\in\{E,k_i,\eta_s,\eta_b\},
  \label{eq:method-positive-parameterization}
\end{equation}
where \(p_{\mathrm{r}}\) is the corresponding reference value and
\(\theta_p\) is the optimized scalar variable for \(p\).  These variables are
combined with prescribed admissible bounds.  

To reduce the number of unknown boundary parameters, the Robin stiffness field can be represented using radial basis functions (RBFs) centered at \(N_c\) prescribed control points \(\bm c_a\), \(a=1,\ldots,N_c\). The stiffness at each Robin support \(i\) is expressed in terms of the unknown coefficients \(\beta_a\) as
\begin{equation}
  k_i
  =
  k_{\mathrm{ref}}
  \exp\!\left(\sum_{a=1}^{N_c}B_{ia}\beta_a\right),
  \qquad
  B_{ia}
  =
  \frac{\varphi(\|\bm{X}_i-\bm{c}_a\|/\ell_R)}
  {\sum_{b=1}^{N_c}\varphi(\|\bm{X}_i-\bm{c}_b\|/\ell_R)} ,
  \label{eq:method-rbf-basis}
\end{equation}
where \(k_{\mathrm{ref}}>0\) is a reference stiffness, \(\beta_a\) are
dimensionless coefficients, and \(B_{ia}\) are normalized basis values. The Gaussian basis \(\varphi(s)=\exp(-s^2/2)\) has length scale \(\ell_R>0\).  The exponential parameterization defines a positive stiffness field using \(N_c\) coefficients, with \(N_c\) smaller than the number of unknown Robin stiffness values.

Let
\(\mathcal{O}\) be the set of observed images.  The simulated image
\(\mathcal{Y}(t;\bm{\theta})\) is compared with the observed image
\(\mathcal{Y}^{\mathrm{o}}(t)\) at each selected time \(t\).  The
identification objective is defined as
\begin{equation}
  \mathcal{J}(\bm{\theta})
  =
  \underbrace{
  \frac{1}{|\mathcal{O}|}
  \sum_{t\in\mathcal{O}}
  \frac{w_t}{3HW}
  \left\|
    \mathcal{Y}(t;\bm{\theta})
    -
    \mathcal{Y}^{\mathrm{o}}(t)
  \right\|_2^2
  }_{\mathcal{J}_{\mathrm{o}}(\bm{\theta})}
  +
  \mathcal{R}(\bm{\theta}),
  \label{eq:method-image-objective}
\end{equation}
where \(H\) and \(W\) are the image height and width in pixels, \(w_t>0\) is the observation weight at time \(t\), and the factor \(3\)
accounts for the image color channels.  
Terminal observations
are used for material and boundary identification near static equilibrium,
whereas transient frames are required to observe the Kelvin--Voigt parameters.
For Robin-field identification, a
Tikhonov prior on logarithmic stiffness is used:
\begin{equation}
  \mathcal{R}(\bm{\theta})
  =
  \frac{\lambda_R}{2N_R}
  \sum_{i=1}^{N_R}
  \left(\log\frac{k_i}{k_i^{\mathrm{prior}}}\right)^2 ,
  \label{eq:method-inverse-regularization}
\end{equation}
where \(N_R\) is the number of Robin supports,
\(k_i^{\mathrm{prior}}>0\) is the prior stiffness, and \(\lambda_R\geq0\)
controls the prior weight.  The penalty is averaged over the support
stiffnesses for both direct and RBF parameterizations.
The estimated parameter vector is then obtained from the bound-constrained
optimization problem
\begin{equation}
  \bm{\theta}^{\star}
  =
  \arg\min_{\bm{\theta}\in\Theta}
  \mathcal{J}(\bm{\theta}),
  \label{eq:method-inverse-argmin}
\end{equation}
where \(\Theta\) denotes the admissible parameter set defined by the prescribed
lower and upper bounds.  In the present implementation,
Eq.~\eqref{eq:method-inverse-argmin} is solved by L-BFGS-B using the gradient
of \(\mathcal{J}\).

This formulation extends inverse analyses in elasticity that identify
Young's modulus and boundary conditions from mechanical observations
\cite{xu2022fembased,xu2023multiple}.  The objective retains the usual structure
of an observation misfit augmented by regularization, while replacing the
displacement residual with an image-domain residual
generated by the coupled 3DGS-SPH forward model.

The gradient required by Eq.~\eqref{eq:method-inverse-argmin} is obtained by
differentiating the image objective subject to the discrete mechanical
equations \(\mathcal Q(\bm z,\bm\theta)=\bm0\) in
Eq.~\eqref{eq:method-forward-residual}, with prescribed initial conditions
independent of \(\bm\theta\).  For fixed reference
appearance, camera, and transfer data, the simulated images depend on the
physical parameters through \(\bm z\).  Their contribution to the objective
sensitivity is
\begin{equation}
  \nabla_{\bm z}\mathcal J_{\mathrm{o}}
  =\sum_{t\in\mathcal O}\frac{2w_t}{3HW|\mathcal O|}
  \left(\frac{\partial\mathcal Y(t)}{\partial\bm z}\right)^T
  \left[\mathcal Y(t)-\mathcal Y^{\mathrm{o}}(t)\right].
  \label{eq:method-trajectory-objective-gradient}
\end{equation}
Here image entries are arranged as vectors when taking derivatives, and
\(\mathcal Y(t)\) denotes the image evaluated from the simulated trajectory.
The image Jacobian includes Gaussian rendering and the Lagrangian SPH--GS
transfer; its transpose maps the image residual to SPH position sensitivities,
as described in Eq.~\eqref{eq:method-map-adjoint-operator}.

Differentiating Eq.~\eqref{eq:method-forward-residual} and eliminating the
trajectory derivative through the adjoint variable \(\bm\lambda\) gives
\begin{equation}
  \left(\frac{\partial\mathcal Q}{\partial\bm z}\right)^T\bm\lambda
  =\nabla_{\bm z}\mathcal J_{\mathrm{o}},
  \qquad
  \nabla_{\bm\theta}\mathcal J
  =\nabla_{\bm\theta}\mathcal R
  -\left(\frac{\partial\mathcal Q}{\partial\bm\theta}\right)^T\bm\lambda.
  \label{eq:method-objective-adjoint-gradient}
\end{equation}
The partial derivatives hold the other argument of \(\mathcal Q\) fixed, whereas
\(\nabla_{\bm\theta}\mathcal J\) is the total gradient along the simulated
trajectory.  Thus the image residual supplies the adjoint forcing, and the
parameter dependence of the mechanical equations determines the observation
contribution to the optimization gradient.

The gradient in Eq.~\eqref{eq:method-objective-adjoint-gradient} is evaluated
by reverse-mode AD implemented in JAX through the discrete trajectory.
This includes implicit differentiation of the
velocity solves, since velocities determine the position increments and
therefore transmit sensitivities across time steps.  The underlying time
integration and matrix-free solver are detailed in
\ref{sec:method-time-integration}.


\section{Results}

The experiments are designed to evaluate the proposed 3DGS-SPH model for
forward simulation and inverse analysis.  Section~\ref{sec:exp10-forward-beam10k}
examines deformation under different constitutive models, boundary conditions,
and object geometries, and Section~\ref{sec:exp30-inverse-recovery} evaluates
the identification of material and boundary parameters from deformation
observations.  Numerical verification of the individual model components is
presented in Appendix.~\ref{app:numerical-verification}.  These tests demonstrate the
accuracy of the SPH displacement response, the computational efficiency and
rendering agreement achieved with the Lagrangian SPH-GS transfer, and the
agreement of AD sensitivities with finite-difference estimates.

SPH simulation, Lagrangian transfer, differentiable Gaussian rendering, and
inverse optimization were conducted on a MacBook Pro equipped with an Apple
M5 Max processor (18 CPU cores and a 40-core integrated GPU) and 48~GB of
unified memory.  The 3DGS reconstructions were trained separately on a laptop equipped with an
NVIDIA GeForce RTX~3050~Ti Laptop GPU.

\subsection{Forward simulation across materials and geometries}
\label{sec:exp10-forward-beam10k}

The forward experiments evaluate the proposed 3DGS-SPH model on a beam, a
bridge, and a liver-shaped object.  The regular beam,  whose surface carries digital image correlation (DIC)-style speckle texture and word ``HKUST'', provides a controlled
benchmark in which the influence of material and boundary settings can be
examined directly.  The bridge and liver-shaped cases then
test the performance of the proposed model for geometrically distinct
reconstructions which are close to real-world applications.  
The design of the forward experiments is summarized in Table~\ref{tab:exp10-forward-design}.  

\begin{table}[!htbp]
  \centering
  \small
  \caption{Design of the forward mechanics experiments.}
  \label{tab:exp10-forward-design}
  \begin{tabular}{p{0.14\linewidth} p{0.16\linewidth} p{0.31\linewidth} p{0.29\linewidth}}
    \toprule
    Geometry & Sim. type & Material model & B.C. setting \\
    \midrule
    Beam & Quasi-static &
    Corotated linear &
    Robin: weak, strong, heterogeneous \\
    Beam & Quasi-static &
    neo-Hookean &
    Robin: weak, heterogeneous \\
    Beam & Dynamic &
    Corotated linear + Kelvin--Voigt &
    Weak Robin \\
    Beam & Dynamic &
    neo-Hookean + Kelvin--Voigt &
    Weak Robin \\
    Bridge & Quasi-static &
    Corotated linear &
    Fixed lower band and central top load \\
    Liver-shaped & Dynamic &
    Corotated linear + Kelvin--Voigt &
    Fixed lower region, Robin region, and end load \\
    \bottomrule
  \end{tabular}
\end{table}
\FloatBarrier

\begin{figure}[!htbp]
  \centering
  \includegraphics[width=0.98\linewidth]{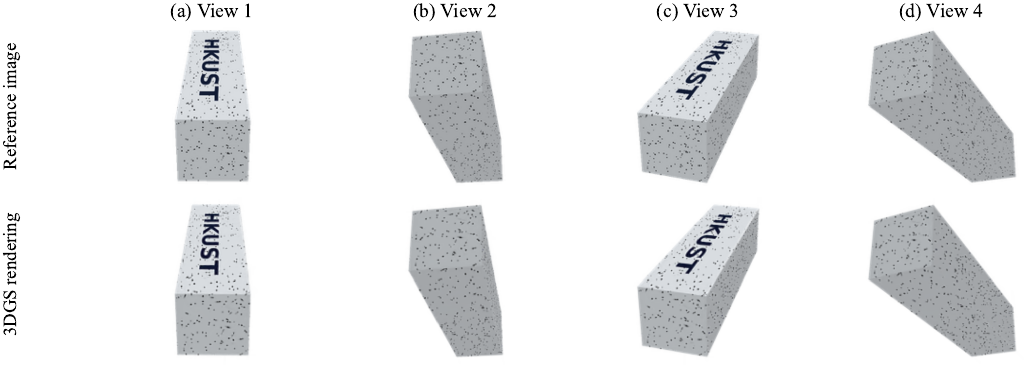}
  \caption{Multiview reconstruction of the undeformed speckle-textured beam
  used in the forward simulations.  The upper row shows four reference
  images from \(\mathfrak D_{\mathrm r}\), and the lower row shows the
  corresponding renderings generated by the reconstructed 10,000-kernel
  3DGS model.}
  \label{fig:exp00b-reconstruction}
\end{figure}

For the beam benchmark, the 10,000 kernel 3DGS model is reconstructed from
multiview images of the undeformed beam.  Figure~\ref{fig:exp00b-reconstruction}
compares reference images in the reference data $\mathfrak{D}_r$ with the corresponding renderings generated by the reconstructed 3DGS model.
The 1225-particle SPH model is generated from the reconstructed geometry using
the procedure in Section~\ref{sec:method-envelope-filling}.

\FloatBarrier

\begin{figure}[!htbp]
  \centering
  \includegraphics[width=0.8\linewidth]{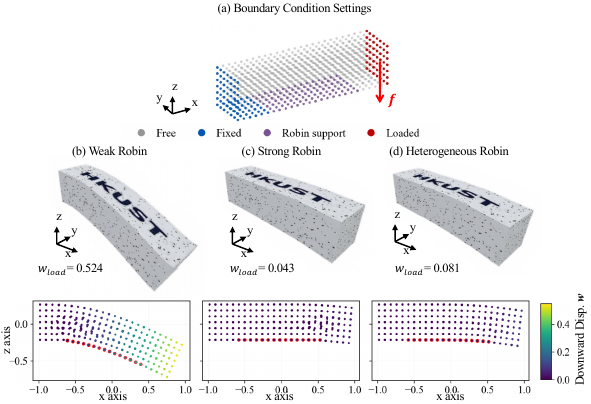}
  \caption{Quasi-static corotated-linear beam response under different Robin
  support fields. Subfigure (a) show the boundary condition settings of the beam case. Subfigures (b)--(d) compare weak, strong, and heterogeneous Robin
  support, respectively. The upper row shows the rendering images of final deformation, and the lower row shows the corresponding SPH particle deformation
  colored by downward displacement \(w=-u_z\).}
  \label{fig:exp10-linear-robin}
\end{figure}

All beam cases use \(E=10^4\,\mathrm{Pa}\), \(\nu=0.35\), \(\rho_0=1200\,\mathrm{kg}\cdot\mathrm{m^{-3}}\), a total downward
load magnitude of \(20\,\mathrm{N}\) on the right end.  
 The left end and the lower surface over the first \(20\%\)
of the beam length are fixed; the designated remaining lower region has
Robin support, and the other unloaded surfaces are free
(Fig.~\ref{fig:exp10-linear-robin}(a)).  The simulations use 1000 steps
with \(\Delta t=0.02\,\mathrm{s}\); numerical velocity damping is set to
\(\gamma_{\mathrm{num}}=2\times10^{-2}\) for quasi-static simulation and
disabled for the dynamic cases.

\FloatBarrier

The quasi-static cases examine the influence of the Robin stiffness field:
weak support with
\(k_R=10^{-2}\,\mathrm{s^{-2}}\), a near-fixed limit
\(k_R=10^4\,\mathrm{s^{-2}}\), and a heterogeneous field
\(k_R=(10^4,10^2)\,\mathrm{s^{-2}}\) over two axial subregions of the same
Robin region, separated at its midpoint.
Figure~\ref{fig:exp10-linear-robin} compares the rendering image of final deformation and
corresponding SPH deformations for the corotated linear elastic model.  Weak restraint
permits substantial bending, whereas strong restraint suppresses motion in
the Robin region; heterogeneous stiffness produces spatially selective
restraint.  Table~\ref{tab:exp10-static-forward} reports the mean downward
displacements of the loaded and Robin regions, denoted by
\(w_{\mathrm{load}}\) and \(w_R\), respectively, and shows that the
neo-Hookean model yields similar responses under the tested conditions.

\begin{table}[!htbp]
  \centering
  \small
  \caption{Quasi-static forward response for the beam.  Downward
  displacement is \(w=-u_z\).}
  \label{tab:exp10-static-forward}
  \begin{tabular}{l l l r r}
    \toprule
    Elastic model & Robin support & \(k_R\) (\(\mathrm{s^{-2}}\)) &
    Final \(w_{\mathrm{load}}\) (m) &
    Final \(w_R\) (m) \\
    \midrule
    corotated linear & weak & \(10^{-2}\) &
    0.5244 & 0.1478 \\
    corotated linear & strong & \(10^4\) &
    0.04265 & \(3.21\times10^{-5}\) \\
    corotated linear & heterogeneous & \(10^4,10^2\) &
    0.08104 & \(5.03\times10^{-3}\) \\
    neo-Hookean & heterogeneous & \(10^4,10^2\) &
    0.08101 & \(5.03\times10^{-3}\) \\
    neo-Hookean & weak & \(10^{-2}\) &
    0.5245 & 0.1475 \\
    \bottomrule
  \end{tabular}
\end{table}

\begin{figure}[!htbp]
  \centering
  \includegraphics[width=0.6\linewidth]{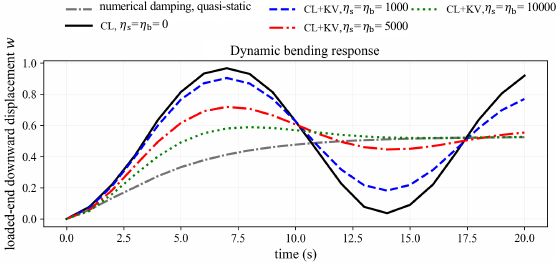}
  \caption{Dynamic bending response of the beam for different viscous coefficients.  The gray curve is the numerically
  damped quasi-static reference; CL denotes the corotated-linear model and KV
  denotes Kelvin--Voigt viscosity.  Increasing \(\eta_s=\eta_b\) reduces the oscillation amplitude.  
  }
  \label{fig:exp10-dynamic-response}
\end{figure}

The dynamic cases retain weak Robin support boundary condition settings and examine Kelvin--Voigt viscosity with \(\eta_s=\eta_b\in\{1000,5000,10000\}\,\mathrm{Pa}\cdot\mathrm{s}\),
alongside an elastic reference with \(\eta_s=\eta_b=0\).
Figure~\ref{fig:exp10-dynamic-response}(a) shows that increasing viscosity
reduces the oscillation amplitude: the peak loaded-end displacement decreases
from \(1.0369\,\mathrm{m}\) in the elastic reference to
\(0.6173\,\mathrm{m}\) at the largest tested viscosity.
The corotated-linear and neo-Hookean responses remain close at
\(\eta_s=\eta_b=5000\,\mathrm{Pa}\cdot\mathrm{s}\)
(Fig.~\ref{fig:exp10-dynamic-response}(b,c)).
Table~\ref{tab:exp10-dynamic-forward} summarizes the final and peak
displacements and SPH simulation times.  The Kelvin--Voigt cases require
more computation than the elastic reference, with runtime increasing with
viscosity over the tested range.

\begin{table}[H]
  \centering
  \small
  \caption{Dynamic forward response with weak Robin support.}
  \label{tab:exp10-dynamic-forward}
  \resizebox{\linewidth}{!}{%
  \begin{tabular}{l l r r r r}
    \toprule
    Elastic model & Viscous model &
    Final \(w_{\mathrm{load}}\) (m) &
    Peak \(w_{\mathrm{load}}\) (m) &
    Final \(\max\|\uvec\|\) (m) &
    time (s) \\
    \midrule
    corotated linear & none &
    0.9188 & 1.0369 & 0.9916 & 11.15 \\
    corotated linear & KV, \(\eta_s=\eta_b=1000\,\mathrm{Pa}\cdot\mathrm{s}\) &
    0.7701 & 0.9677 & 0.8186 & 37.83 \\
    corotated linear & KV, \(\eta_s=\eta_b=5000\,\mathrm{Pa}\cdot\mathrm{s}\) &
    0.5540 & 0.7602 & 0.5794 & 71.11 \\
    corotated linear & KV, \(\eta_s=\eta_b=10000\,\mathrm{Pa}\cdot\mathrm{s}\) &
    0.5267 & 0.6173 & 0.5501 & 92.68 \\
    neo-Hookean & KV, \(\eta_s=\eta_b=5000\,\mathrm{Pa}\cdot\mathrm{s}\) &
    0.5548 & 0.7628 & 0.5804 & 72.55 \\
    \bottomrule
  \end{tabular}}
\end{table}
\FloatBarrier


\begin{figure}[!htbp]
  \centering
  \includegraphics[width=0.7\linewidth]{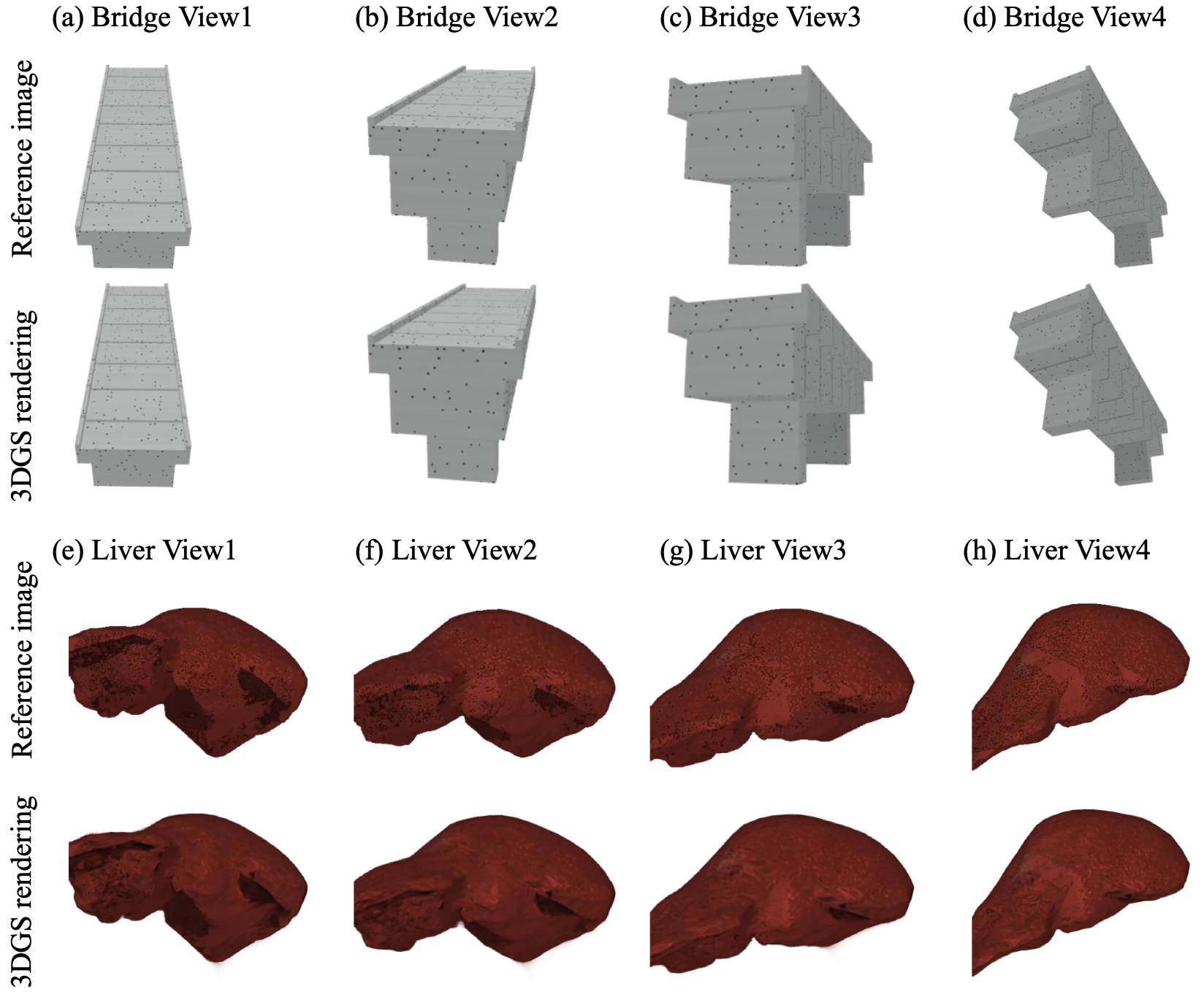}
  \caption{Multiview reconstruction of the undeformed bridge (a--d) and
  liver-shaped object (e--h).  For each object, the upper row shows reference
  images from its dataset \(\mathfrak D_{\mathrm r}\), and the lower row
  shows the corresponding renderings generated by the reconstructed 3DGS
  model.}
  \label{fig:exp10-cross-geometry-reconstruction}
\end{figure}
\FloatBarrier

\begin{figure}[!htbp]
  \centering
  \includegraphics[width=0.9\linewidth]{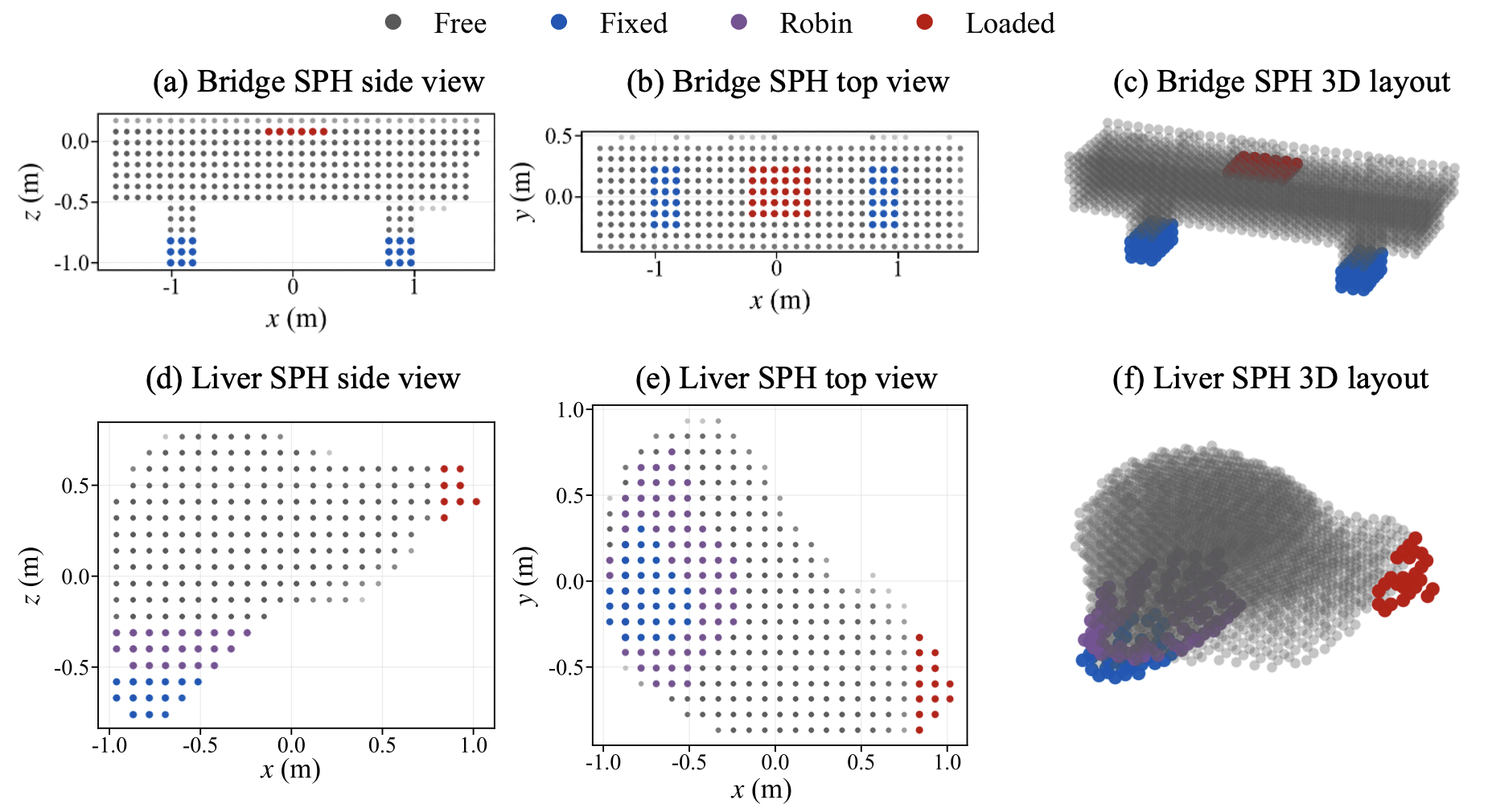}
  \caption{Boundary configurations for the bridge and liver-shaped forward
  simulations.   The bridge case uses a lower fixed band and a central loaded
  patch on the top layer. The liver-shaped case uses surface supports for the
  lower fixed region, the adjacent Robin region, and the loaded end region.}
  \label{fig:exp10-cross-geometry-boundaries}
\end{figure}

The bridge and liver-shaped cases extend the forward simulations to
irregular geometries reconstructed from reference images in
\(\mathfrak D_{\mathrm r}\)
(Fig.~\ref{fig:exp10-cross-geometry-reconstruction}).  Both use
corotated-linear elasticity, with Kelvin--Voigt viscosity added for the
liver-shaped case.  Their material parameters and load magnitudes are
listed in Table~\ref{tab:exp10-cross-geometry-parameters}.
The simulations use 1000 steps with \(\Delta t=0.02\,\mathrm{s}\), with numerical
velocity damping \(\gamma_{\mathrm{num}}=0.02\) for the quasi-static bridge
and zero for the dynamic liver-shaped object.  Their boundary conditions
are shown in Fig.~\ref{fig:exp10-cross-geometry-boundaries}.
The bridge is fixed over the lowest 20\% of its height and loaded downward
on the uppermost particle layer within a central square of side length
equal to half its width.  In contrast, the
liver-shaped object uses surface supports, fixed over the lowest 15\% of
its height and restrained by \(k_R=100\,\mathrm{s^{-2}}\) in the adjacent
15--30\% band.  Its downward load acts within the final 10\% of its extent
along the positive \(x\)-direction, excluding the fixed and Robin regions.

\begin{table}[!htbp]
  \centering
  \caption{Material parameters and loading for the bridge and liver-shaped
  forward simulations.}
  \label{tab:exp10-cross-geometry-parameters}
  \begin{tabular}{l c c c c c}
    \toprule
    Geometry & \(E\) (Pa) & \(\nu\) &
    \(\rho_0\) (\(\mathrm{kg}\cdot\mathrm{m^{-3}}\)) &
    \(\eta_s=\eta_b\) (\(\mathrm{Pa}\cdot\mathrm{s}\)) & Load (N) \\
    \midrule
    Bridge & \(10^5\) & 0.20 & 2400 & 0 & 500 \\
    Liver-shaped & \(10^4\) & 0.45 & 1060 & 3000 & 8 \\
    \bottomrule
  \end{tabular}
\end{table}

\FloatBarrier

\begin{figure}[!htbp]
  \centering
  \includegraphics[width=0.7\linewidth]{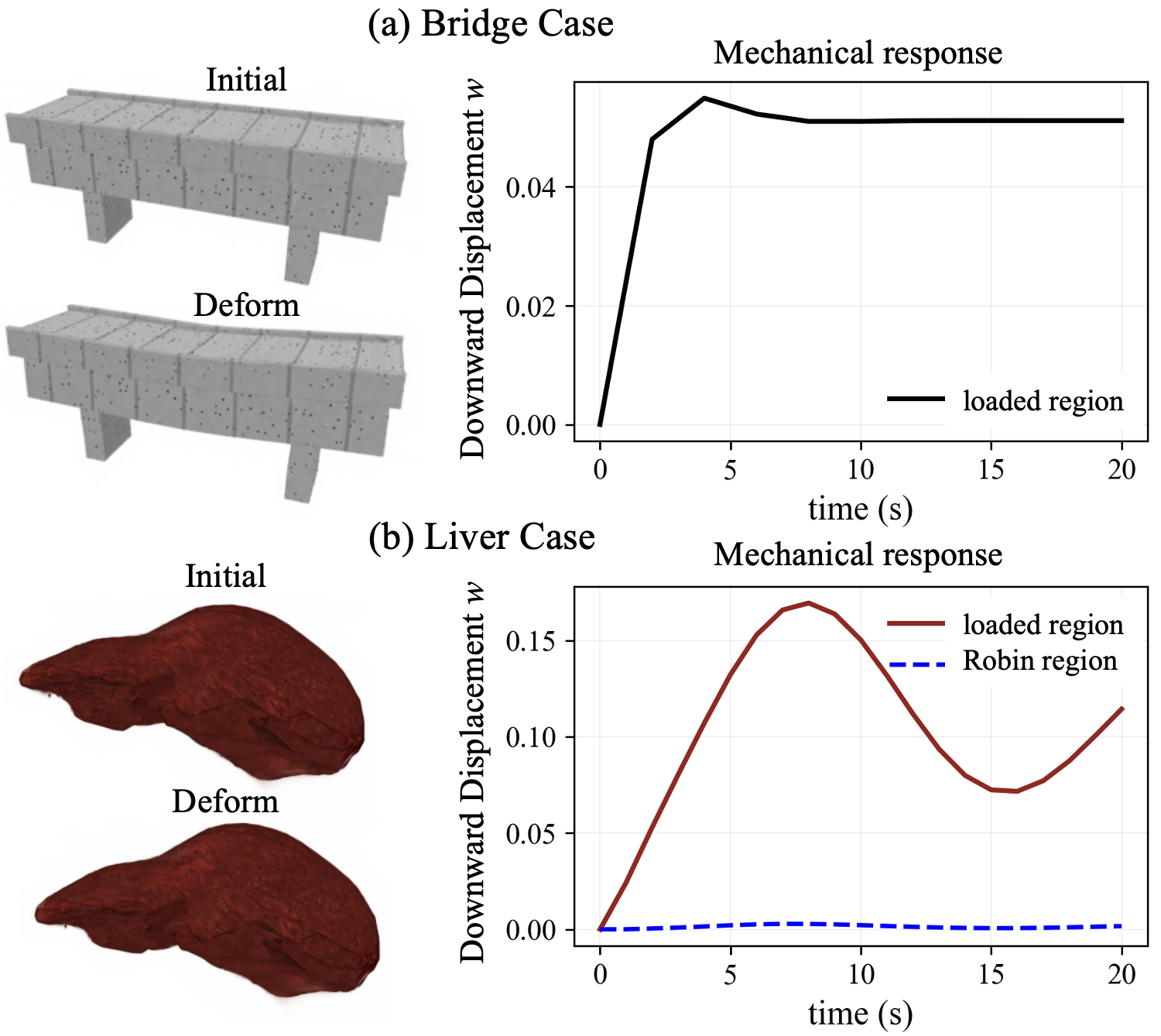}
  \caption{Forward deformation responses for the bridge and liver-shaped
  examples.  Initial and deformed Gaussian renderings are shown together with
  the corresponding mean downward displacement histories.  (a) The bridge case
  illustrates a quasi-static elastic response on a multi-member structure.
  (b) The liver-shaped case illustrates a transient viscoelastic response
  with both loaded-region and Robin-region displacement curves.}
  \label{fig:exp10-cross-geometry}
\end{figure}

Under these conditions, the bridge
approaches a stationary deflected configuration, whereas the liver-shaped
object exhibits a damped transient response in the loaded and Robin regions.
Figure~\ref{fig:exp10-cross-geometry} presents these responses through the
initial and deformed Gaussian renderings and displacement histories.
Table~\ref{tab:exp10-cross-geometry} summarizes the final and peak
loaded-region displacements and SPH simulation times alongside the model
sizes.  Together with the beam results, these
cases demonstrate quasi-static and dynamic forward simulation using the
3DGS-SPH model across distinct geometries, constitutive models, and boundary
conditions.

\begin{table}[H]
  \centering
  \small
  \caption{Forward simulation results for the bridge and liver-shaped
  cases.  Final and peak \(w_{\mathrm{load}}\)
  denote the mean loaded-region displacement at \(t=20\,\mathrm{s}\) and
  its maximum over the simulated interval.}
  \label{tab:exp10-cross-geometry}
  \setlength{\tabcolsep}{4pt}
  \begin{tabular}{l r r r r r}
    \toprule
    Geometry & \shortstack{SPH particles} & Gaussians &
    \shortstack{Final \(w_{\mathrm{load}}\) (m)} &
    \shortstack{Peak \(w_{\mathrm{load}}\) (m)} &
    \shortstack{time (s)} \\
    \midrule
    Bridge & 2381 & 11310 & 0.05116 & 0.06392 & 32.85 \\
    Liver-shaped & 2677 & 10000 & 0.11434 & 0.19272 & 123.28 \\
    \bottomrule
  \end{tabular}
\end{table}
\FloatBarrier


\FloatBarrier

\subsection{Inverse analysis from rendered deformation observations}
\label{sec:exp30-inverse-recovery}

The inverse analysis experiments evaluate material and boundary parameter
identification from images of deformation observations using the proposed
3DGS-SPH model.  The beam provides the principal benchmark, while the bridge
and liver-shaped examples examine selected inverse capabilities on distinct
geometries.  All cases minimize the image-mismatch objective in
Eq.~\eqref{eq:method-image-objective} using bound-constrained L-BFGS-B and
AD gradients, with case-specific regularization described below.
Gradient verification is reported in \ref{sec:exp20-ad-validation}.

The experiment design is summarized in Table~\ref{tab:exp30-inverse-design}.
The cases progress from separate material and boundary identification to
joint recovery and hierarchical static-dynamic identification.
All inverse simulations use 1000 steps with \(\Delta t=0.02\,\mathrm{s}\).
Quasi-static cases use numerical velocity damping
\(\gamma_{\mathrm{num}}=2\times10^{-2}\) and terminal observations;
dynamic cases disable numerical damping and use five transient images
selected from states saved every 100 steps.

\begin{table}[!htbp]
  \centering
  \small
  \caption{Design of the inverse-identification experiments.}
  \label{tab:exp30-inverse-design}
  \begin{tabular}{p{0.17\linewidth} p{0.32\linewidth} p{0.39\linewidth}}
    \toprule
    Geometry & Unknown parameters & Material model \\
    \midrule
    Beam & \(E\) & corotated linear \\
    Bridge & \(E\) & neo-Hookean \\
    Beam & \(E,\eta_s,\eta_b\) & corotated linear + Kelvin--Voigt \\
    Beam & \(k_R\) & corotated linear \\
    Beam & \(E,k_R\) & corotated linear / neo-Hookean \\
    Liver-shaped & \(E,k_R\), then \(\eta_s,\eta_b\) & neo-Hookean + Kelvin--Voigt \\
    \bottomrule
  \end{tabular}
\end{table}
\FloatBarrier

\subsubsection{Material parameters identification under prescribed boundary conditions}

\begin{figure}[!htbp]
  \centering
  \includegraphics[width=0.9\linewidth]{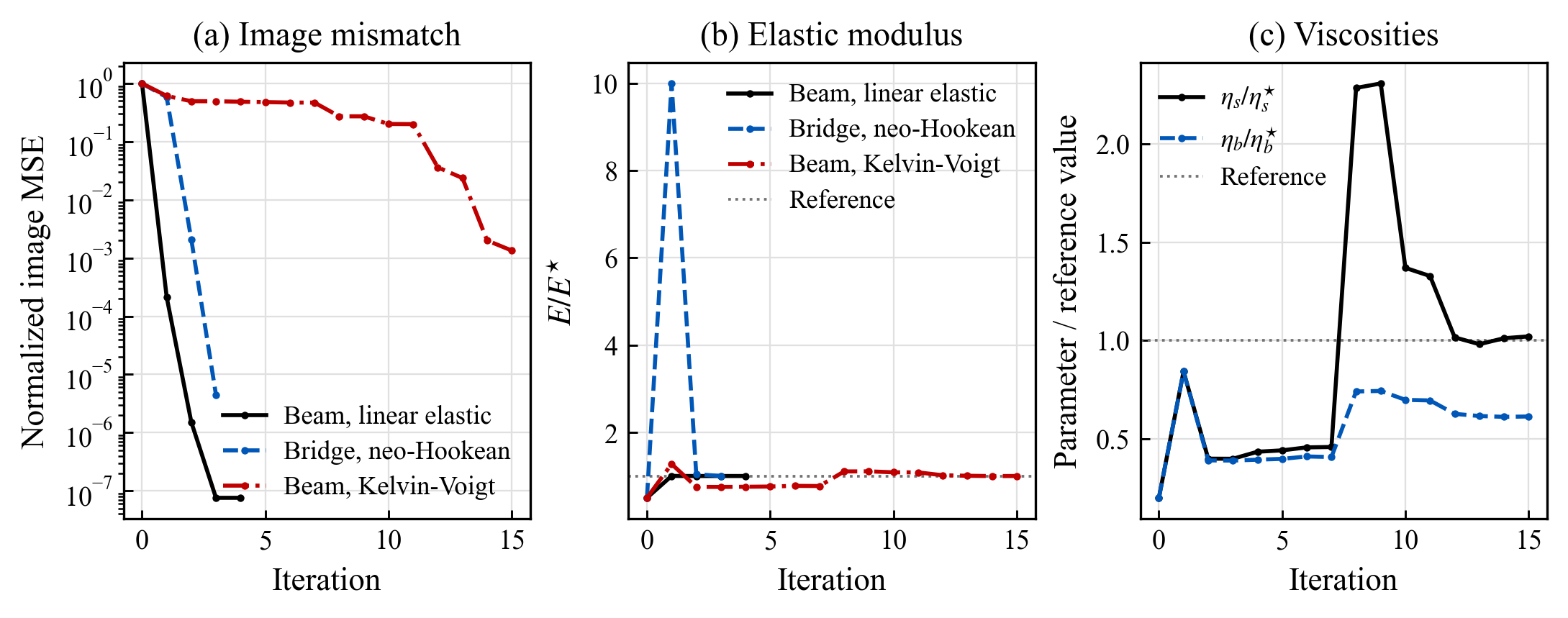}
  \caption{Iteration histories for constitutive identification under
  prescribed boundary conditions. (a) Image-data mismatch objective \(\mathcal J_{\mathrm{o}}\) normalized by its
  initial value. (b) Elastic modulus relative to its reference. (c) Recovered
  Kelvin--Voigt coefficients in the five-frame beam test. Horizontal dotted
  lines denote the reference values.}
  \label{fig:exp30-constitutive-convergence}
\end{figure}

With boundary conditions prescribed, constitutive parameters are identified
using the image-mismatch term \(\mathcal J_{\mathrm{o}}\) without
regularization.  
The beam retain the weak Robin support and fixed support with the downward end load from the forward case (Fig.~\ref{fig:exp10-linear-robin}(a)).  The bridge retains the lower fixed band and central downward
load of \(500\,\mathrm{N}\) from the forward case
(Fig.~\ref{fig:exp10-cross-geometry-boundaries}), but uses neo-Hookean
elasticity.
Figure~\ref{fig:exp30-constitutive-convergence} compares
the optimization histories for the corotated-linear beam, neo-Hookean
bridge, and Kelvin--Voigt beam.  In the two elastic cases, Young's modulus
is initialized at half its reference value, with
\(E^\star=10^4\,\mathrm{Pa}\) for the beam and
\(10^5\,\mathrm{Pa}\) for the bridge.  The recovered moduli have relative
errors below \(0.001\%\) and \(0.015\%\), respectively, with final image
mean-squared errors (MSEs) below \(3\times10^{-11}\).

\FloatBarrier

\begin{figure}[!htbp]
  \centering
  \includegraphics[width=0.8\linewidth]{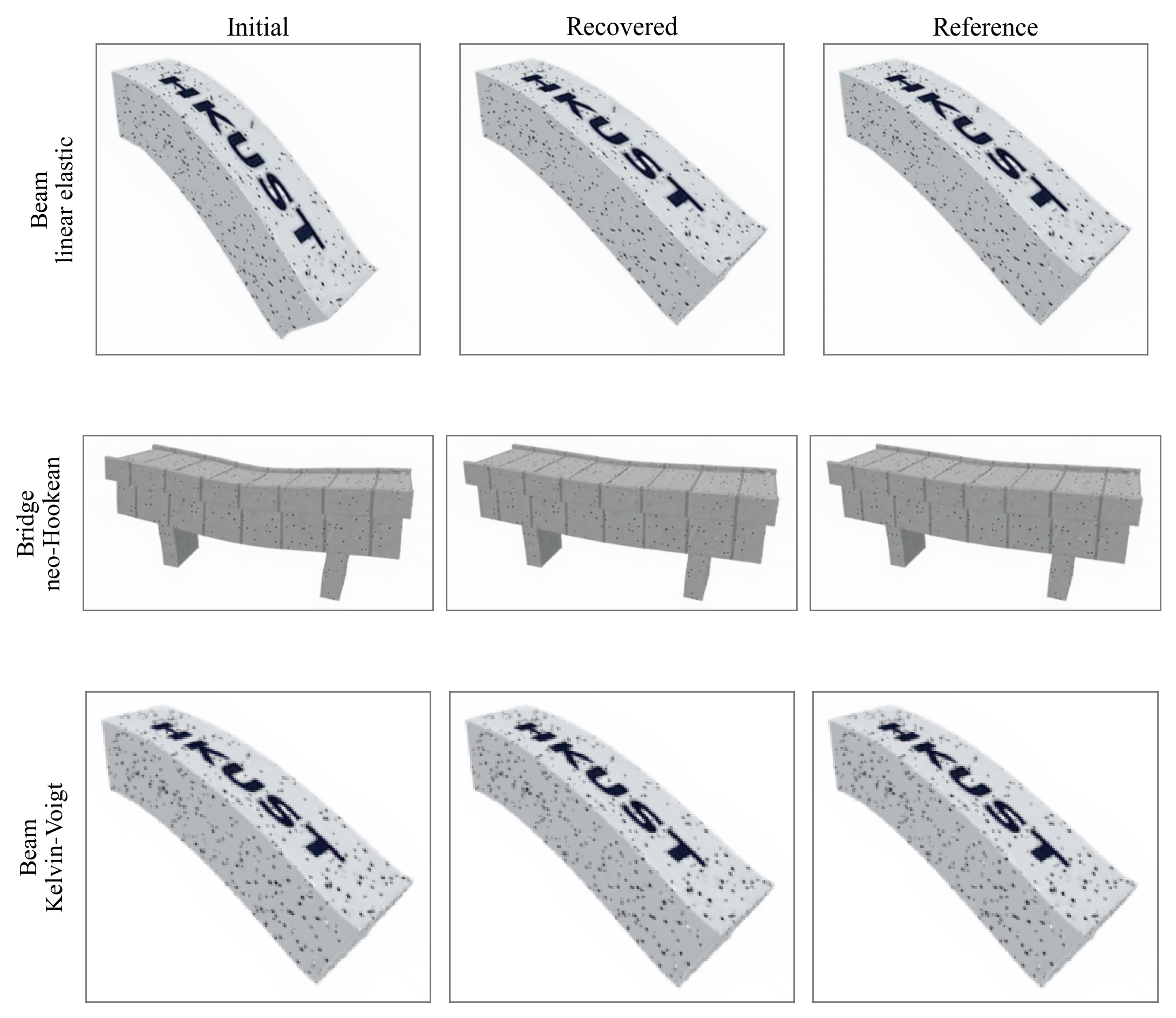}
  \caption{Forward-evaluated states for constitutive identification. Columns
  show the initial estimate, recovered state, and synthetic reference. Rows
  correspond to the corotated linear beam, neo-Hookean bridge, and the last
  observed frame of the Kelvin--Voigt beam.}
  \label{fig:exp30-constitutive-states}
\end{figure}

The dynamic beam estimates \((E,\eta_s,\eta_b)\) under prescribed weak
Robin support \(k_R=10^{-2}\,\mathrm{s^{-2}}\), retaining the load and fixed lower support in the corresponding quasi-static case.
The reference values are
(10000\,$\mathrm{Pa}$,5000\,$\mathrm{Pa}\cdot\mathrm{s}$,5000\,$\mathrm{Pa}\cdot\mathrm{s}$), and
the initial values are
(5000\,$\mathrm{Pa}$,1000\,$\mathrm{Pa}\cdot\mathrm{s}$,
1000\,$\mathrm{Pa}\cdot\mathrm{s}$).
The multi-frame MSE decreases from \(7.74\times10^{-3}\) to
\(1.04\times10^{-5}\); \(E\) and \(\eta_s\) are recovered within
\(0.012\%\) and \(2.14\%\), whereas \(\eta_b\) is underestimated by
\(38.6\%\). 
The larger discrepancy in \(\eta_b\) is consistent with the
bending-dominated loading, which primarily induces distortional rather than
volumetric deformation.  Since bulk viscosity governs the volumetric
strain-rate contribution, the selected images may constrain \(\eta_b\)
only weakly.  Nevertheless, the recovered parameters reproduce the observed
transient response under the tested loading and boundary conditions, with
close agreement between the recovered and reference rendered states
(Fig.~\ref{fig:exp30-constitutive-states}).  More precise identification of
bulk viscosity warrants dedicated loading and observation designs that
better resolve volumetric viscous effects, which are beyond the scope of
the present study.

\FloatBarrier

\subsubsection{Joint identification of material parameters and boundary conditions}

The beam tests jointly estimate Young's modulus \(E\) and the heterogeneous
Robin stiffness field from one terminal image, retaining the quasi-static
loading and boundary regions of the tests in section 4.2.2.
The reference values are
\(E^\star=10^4\,\mathrm{Pa}\)
and \(k_R^\star=(10^4,10^2)\,\mathrm{s^{-2}}\).
The particlewise and RBF representations are compared under corotated-linear
and neo-Hookean elasticity, starting from \(E_0=5000\,\mathrm{Pa}\)
and \(k_{R,0}=10^3\,\mathrm{s^{-2}}\).  The objective combines
\(\mathcal J_{\mathrm{o}}\) with the log-stiffness prior weighted by
\(\lambda_R=10^{-6}\).

\begin{figure}[!htbp]
  \centering
  \includegraphics[width=0.75 \linewidth]{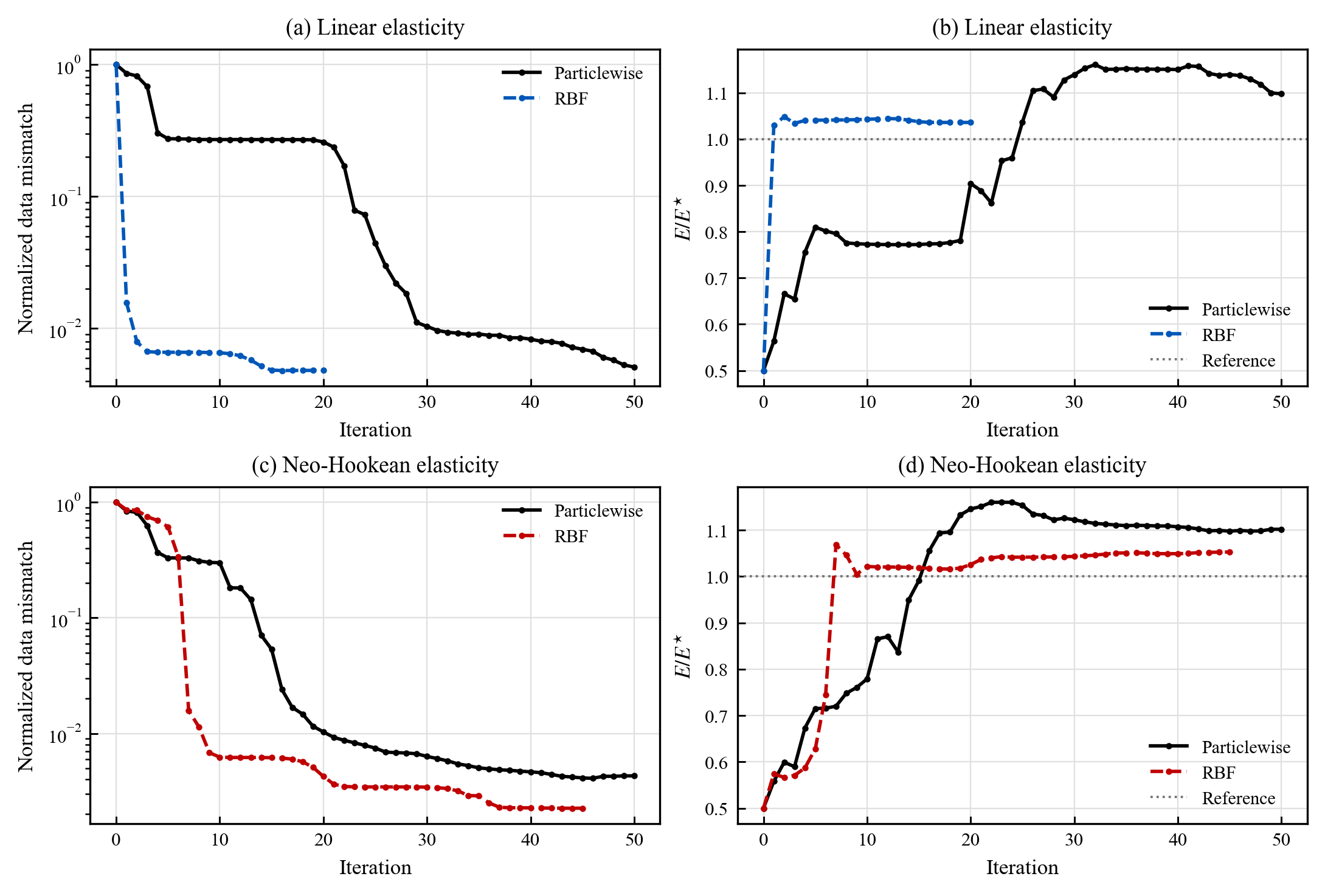}
  \caption{Joint identification of material parameters and boundary conditions
  for the beam. (a,c) Normalized image-data mismatch for the corotated-linear
  and neo-Hookean models. (b,d) Corresponding Young's modulus estimates
  normalized by the reference value. Particlewise and RBF boundary
  representations use the same scene configuration. The corotated-linear
  particlewise history reports best-so-far evaluations.}
  \label{fig:exp30-joint-convergence}
\end{figure}

\begin{figure}[!htbp]
  \centering
  \includegraphics[width=0.8\linewidth]{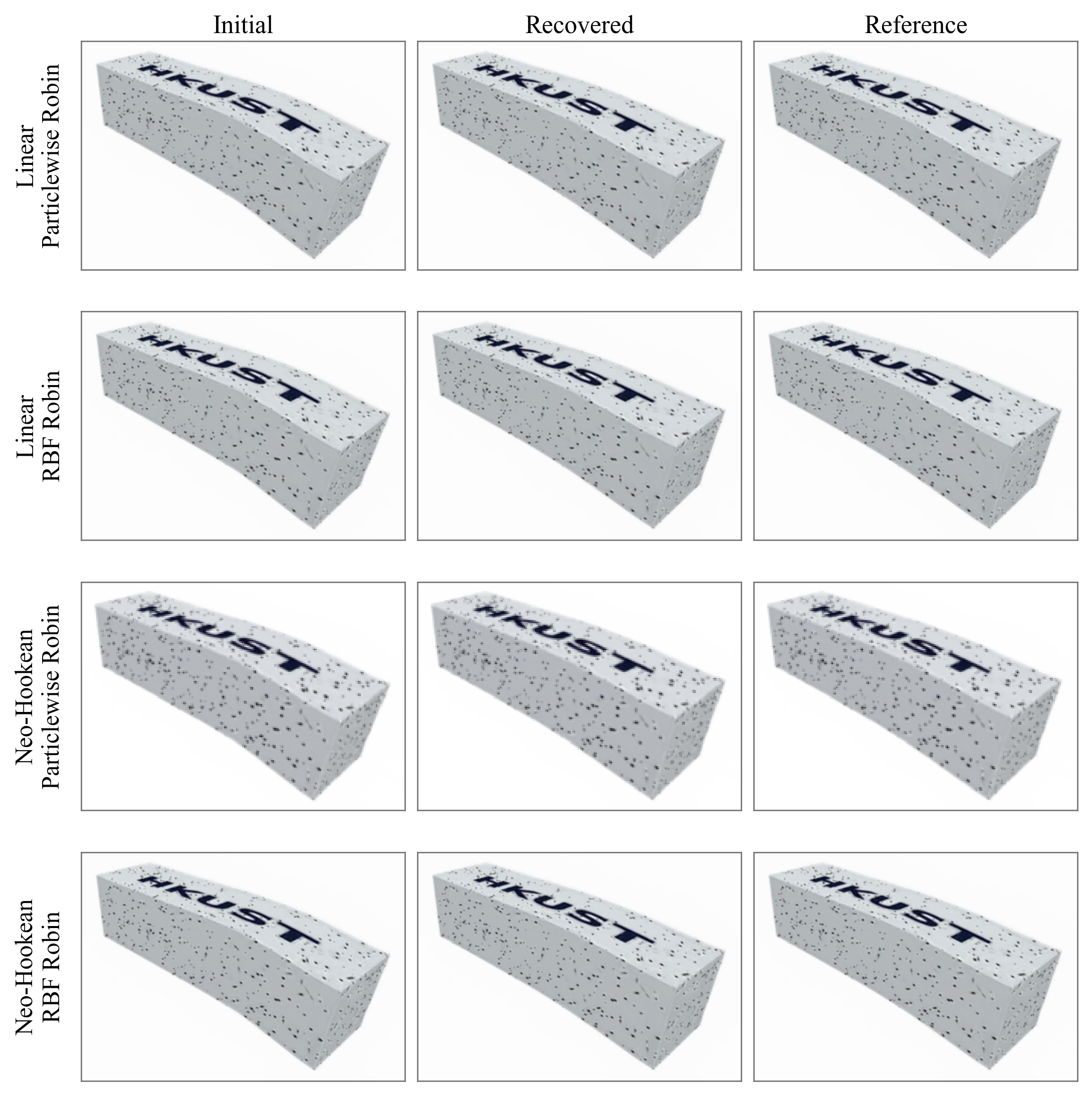}
  \caption{Final rendered states for joint identification of material
  parameters and boundary conditions for the beam. Columns show
  initialization, recovery, and reference. Rows
  compare particlewise and RBF Robin fields under corotated linear and
  neo-Hookean elasticity.}
  \label{fig:exp30-joint-states}
\end{figure}

All four cases reduce the image mismatch
(Fig.~\ref{fig:exp30-joint-convergence}), reaching final MSEs between
\(2.26\times10^{-6}\) and \(6.18\times10^{-6}\).
The RBF representation gives modulus errors of \(3.62\%\) and \(5.27\%\)
for corotated-linear and neo-Hookean elasticity, respectively, compared
with \(9.80\%\) and \(10.19\%\) for the particlewise representation.
Its Robin-region displacement RMSE remains below
\(7.1\times10^{-4}\,\mathrm{m}\) in both cases.
The recovered renderings in Fig.~\ref{fig:exp30-joint-states} reproduce
the reference deformation, demonstrating simultaneous material and
boundary-response recovery.

\FloatBarrier

\begin{figure}[!htbp]
  \centering
  \includegraphics[width=\linewidth]{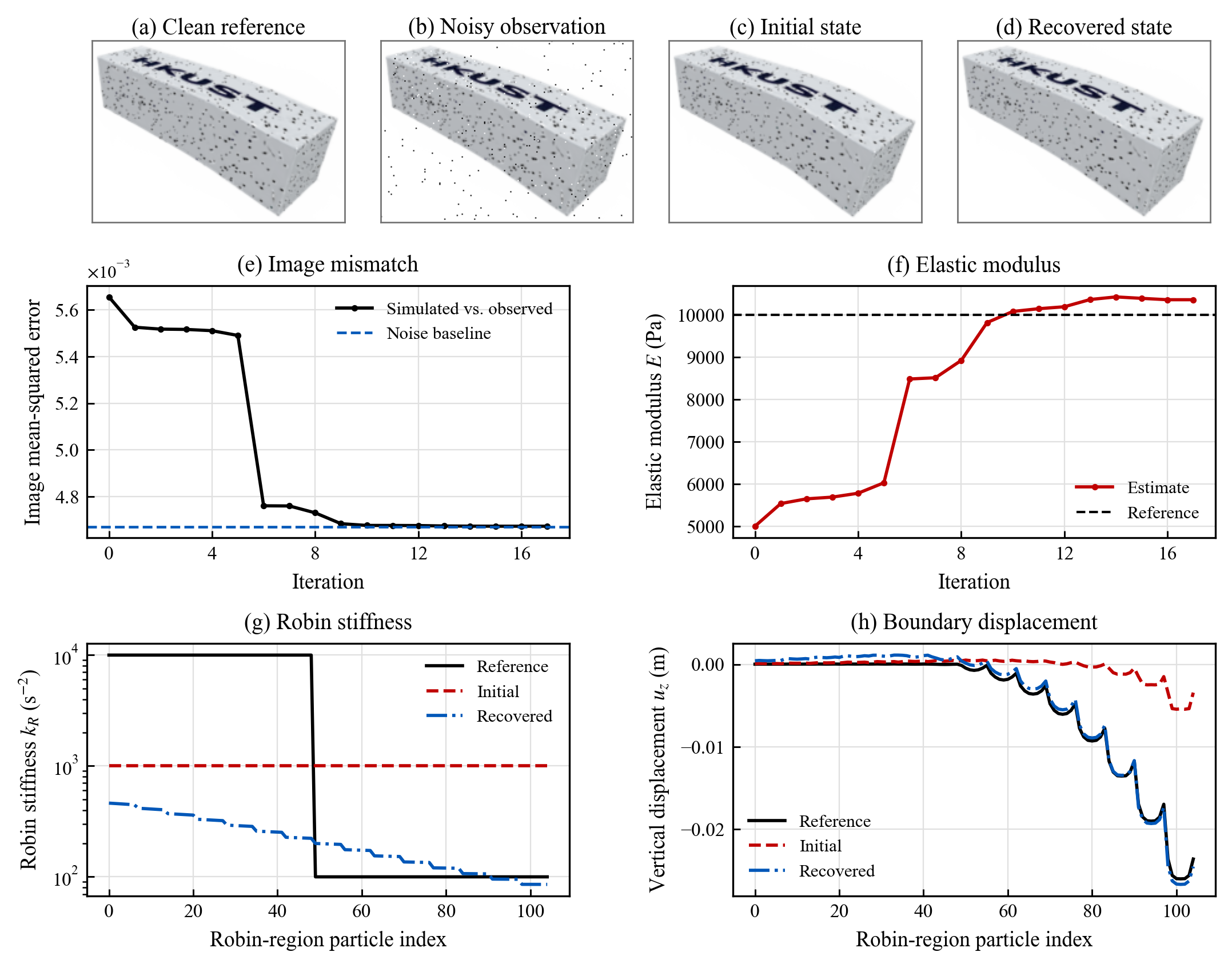}
  \caption{Joint identification of material parameters and boundary conditions
  for the beam using an image with 1\% salt-and-pepper noise.
  (a,b) Clean reference and noisy observation.
  (c,d) Simulated images with the initial and recovered parameters.
  (e) Image MSE between the simulated image and noisy observation during
  optimization; the dashed line denotes the MSE between the clean reference
  and noisy observation.
  (f) Elastic modulus estimates and reference value.
  (g,h) Reference, initial, and recovered Robin stiffness and final vertical
  displacement over the Robin region. Particles in (g,h) are ordered by
  increasing reference \(x\)-coordinate from the fixed-end side to the
  loaded-end side.}
  \label{fig:exp30-salt-pepper-robustness}
\end{figure}

Robustness to observation noise is examined using the corotated-linear beam
and eight-control RBF field.  Each reference-image pixel is independently
replaced by black or white with a total probability of 1\%, with equal
probabilities for the two values.  This realization remains fixed during
optimization; noise is not added to the simulated images.  The initial
parameters remain \(E_0=5000\,\mathrm{Pa}\) and
\(k_{R,0}=10^3\,\mathrm{s^{-2}}\), with prior stiffness
\(1\,\mathrm{s^{-2}}\) and \(\lambda_R=10^{-8}\).
Figure~\ref{fig:exp30-salt-pepper-robustness} shows recovery of the modulus
within \(3.59\%\) of \(E^\star=10^4\,\mathrm{Pa}\), while the
observation MSE approaches the noise floor.  Evaluated against the clean
reference, image MSE decreases from \(9.98\times10^{-4}\) to
\(3.43\times10^{-6}\), and global displacement RMSE decreases from
\(8.45\times10^{-3}\,\mathrm{m}\) to
\(5.49\times10^{-4}\,\mathrm{m}\).  The test therefore demonstrates
effective material and boundary response recovery despite the imposed
image noise.

These results show that the boundary representation influences material identification as well as boundary recovery. Under both constitutive models, the RBF representation yields smaller Youngs modulus errors while reproducing the displacement over the Robin region. This suggests that a compact boundary parameterization helps constrain the joint estimation of material stiffness and boundary restraint from the observed deformation.

Overall, the inverse experiments demonstrate that the proposed 3DGS-SPH model
can be applied seamlessly to inverse analysis of physical parameter estimation from images of deformation observations. Across the beam, bridge, and liver-shaped cases, the model recovers
material parameters, Robin-region displacement responses, and selected
viscoelastic parameters within a unified differentiable inverse-analysis
framework. The salt-and-pepper noise case further shows that effective
material and boundar -response recovery can be maintained when the observed
image contains a representative impulse-noise perturbation.

\FloatBarrier

\section{Discussion}

The beam comparison demonstrates the computational advantage of the proposed
3DGS--SPH model over the co-located construction.  Using Gaussian centers as
mechanical particles ties the SPH discretization to the dense,
surface-concentrated distribution required for rendering, increasing the
mechanical cost without necessarily improving volumetric support quality.
The proposed Lagrangian SPH--GS transfer removes this dependence, allowing
a coarser volumetric SPH model to drive the same Gaussian rendering model.
In the tested beam case, this reduces mechanical simulation time by a factor
of 48 while retaining close rendering agreement at matched deformation
amplitudes (Appendix~\ref{sec:exp00b-lagrangian-map}).

For inverse analysis, the sensitivity verification in
Appendix~\ref{sec:exp20-ad-validation} identifies a controlled approximation:
corotated-linear rotations are updated during forward simulation but held
fixed during differentiation.  Although this produces a larger AD--FD
discrepancy than in the other tested parameters, accurate modulus recovery
in the prescribed-boundary beam case indicates that the resulting gradients
remain effective for that identification task.

The boundary results distinguish recovery of an effective displacement
response from identification of the local stiffness field.  As Robin
restraint approaches free or fixed behavior, deformation becomes less
informative about stiffness magnitude, allowing irregular particlewise
estimates to reproduce the observed response.  The RBF representation
restricts the field to a smoother, lower-dimensional space and improves
recovery when the support variation is compatible with that representation.
Its smaller modulus errors in both joint-identification cases further
indicate that boundary parameterization affects material estimation as well
as support recovery.  Parameterization and regularization therefore help
constrain the admissible solutions, rather than merely reduce image mismatch.

The dynamic cases similarly highlight the role of observation sensitivity.
More accurate recovery of \(\eta_s\) than \(\eta_b\) in the
prescribed-boundary beam test suggests that the bending observations
constrain shear viscosity more strongly than bulk viscosity.  In the
hierarchical analysis, residual errors in the identified modulus and
boundary field can also influence subsequent viscosity estimates.  These
results motivate richer transient excitation and observation sequences to
better distinguish elastic, boundary, and viscous effects within the same
coupled model.

\section{Conclusion}

This work developed a differentiable Lagrangian-coupled 3DGS-SPH model for
forward simulation and inverse analysis of deformable solids reconstructed
from image or video observations.  The central idea is to construct a
mechanics-oriented SPH model
from the reconstructed Gaussian envelope and to couple this physical model to
the Gaussian representation through a reference-configuration Lagrangian
transfer.  In this way, a lower-dimensional SPH model can provide a
mechanically reasonable discretization for constitutive response, boundary
conditions, and time integration, while the higher-resolution Gaussian model
retains its role as the surface geometry and appearance representation for
rendering.  The JAX implementation further preserves the differentiable
structure of the coupled simulation, transfer, and rendering process, enabling
automatic-differentiation gradients for the inverse analysis pipeline.

The numerical studies demonstrate the main capabilities of the proposed
model.  The corrected solid-SPH response was compared with finite-element
reference results, and the Lagrangian SPH-GS transfer was shown to
produce rendered deformation close to the co-located construction while using
a substantially lower-dimensional physical discretization.  Forward examples
on beam, bridge, and liver-shaped objects further show that the same coupled
model can represent linear elastic, hyperelastic, and Kelvin--Voigt
viscoelastic responses under different boundary settings.  The sensitivity
checks and inverse examples verify that the implemented forward process can be
used for gradient-based estimation of constitutive and boundary parameters
from deformation images or video frames.

Overall, the results support the proposed 3DGS-SPH model as a practical
route for connecting reconstructed object appearance with solid-mechanics
simulation and parameter identification.  Future work will extend the same
modeling framework toward contact and interaction problems, inverse analysis
under evolving boundary conditions, geometry-aware SPH-GS transfer, and
more complex constitutive and loading scenarios.

\section*{Acknowledgements}

This work was financially supported by the National Natural Science
Foundation of China (YSF 1240020027, JRS N\_HKUST687/25), the Research Grants
Council of Hong Kong (ECS 26205024, CRF C7085-24G), the Innovation and
Technology Fund (ITS/032/23), and the Office of Knowledge Transfer of
The Hong Kong University of Science and Technology (BGF.2026.019).

\FloatBarrier
\setcounter{figure}{0}
\setcounter{table}{0}
\renewcommand{\theHfigure}{appendix.\Alph{section}.\arabic{figure}}
\renewcommand{\theHtable}{appendix.\Alph{section}.\arabic{table}}

\newpage
\appendix
\begin{center}
    \noindent{\Large\bfseries Appendix}
\end{center}
\section{Corrected Lagrangian SPH formulation and implicit solution}
\label{app:sph-formulation}

This appendix details the physical model summarized in
Section~\ref{sec:method-sph-discretization}.  It specifies the corrected
spatial discretization, constitutive stresses, and matrix-free implicit
operator used by \(\mathcal S_{\mathrm{SPH}}\). Compared with the corrected SPH method proposed by Peer et al.~\cite{peer2018implicit}, we further extend neo-Hookean
hyperelasticity, Kelvin–Voigt viscosity, and Robin-like boundary condition to this implementation.

\subsection{Corrected SPH formulation}
\label{sec:method-corrected-sph}

For the supports and reference graph defined in
Section~\ref{sec:method-sph-discretization}, the SPH approximation of a
scalar or vector field \(A\) is
\begin{equation}
  A(\bm{X}_i)
  \simeq
  \sum_{j\in\mathcal{N}_i}
  A(\bm{X}_j) W_{ij}^{0}\Delta V_j,
  \qquad
  W_{ij}^{0}
  =
  W(\|\bm{X}_i-\bm{X}_j\|,h),
  \label{eq:method-sph-interpolation}
\end{equation}
The superscript \(0\) denotes evaluation in the reference configuration, and
\(W\) is the cubic spline kernel in
Eq.~\eqref{eq:method-cubic-spline-kernel}.  

Direct use of the uncorrected kernel gradient may lose first-order consistency
when the support distribution is irregular or one-sided near boundaries.
The adopted correction follows Peer et al.~\cite{peer2018implicit} and
related corrected-gradient SPH methods~\cite{randles1996sph,bonet1999variational}.
With \(\nabla_i W_{ij}^{0}=\nabla_{\!\bm{X}_i}W_{ij}^{0}\), the
correction matrix is defined as
\begin{equation}
  \bm{L}_i
  =
  \left[
    \sum_{j\in\mathcal{N}_i}
    \Delta V_j
    (\bm{X}_j-\bm{X}_i)\otimes\nabla_i W_{ij}^{0}
  \right]^{-T},
  \qquad
  \widehat{\nabla}_i W_{ij}^{0}
  =
  \bm{L}_i\nabla_i W_{ij}^{0},
  \label{eq:method-gradient-correction}
\end{equation}
This correction enforces reproduction of affine displacement fields through the
discrete gradient operator.  In the total-Lagrangian setting, the neighbor graph
and correction matrices \(\bm{L}_i\) are constructed once in the reference
configuration.  The deformation gradient at support \(i\) is then evaluated as
\begin{equation}
  \bm{F}_i^n
  =
  \sum_{j\in\mathcal{N}_i}
  \Delta V_j
  (\bm{x}_j^n-\bm{x}_i^n)
  \otimes
  \widehat{\nabla}_iW_{ij}^{0}.
  \label{eq:method-sph-F}
\end{equation}
Here \(n\) is the time-step index.  In a corotated formulation, the local
rigid-body rotation is removed before applying a linear strain law, so that
large rotations are not interpreted as artificial elastic strain.  A local
rotation \(\bm{R}_i^n\) is therefore extracted from the corrected deformation
gradient following the robust rotation extraction of
M\"uller et al.~\cite{muller2016rotation}.  The corresponding rotated corrected
gradient is
\begin{equation}
  \widehat{\nabla}_i W_{ij}^{\,n}
  =
  \bm{R}_i^n\widehat{\nabla}_i W_{ij}^{0},
  \label{eq:method-rotated-corrected-gradient}
\end{equation}
which enters the corotated stress-divergence operator.  The corotated
deformation measure is evaluated from the relative displacement between the
current neighbor vector and the rotated reference neighbor vector:
\begin{equation}
  \bm{F}_{i}^{\mathrm{co},n}
  =
  \bm{I}
  +
  \sum_{j\in\mathcal{N}_i}
  \Delta V_j
  \left[
    (\bm{x}_j^n-\bm{x}_i^n)
    -
    \bm{R}_i^n(\bm{X}_j-\bm{X}_i)
  \right]
  \otimes
  \widehat{\nabla}_i W_{ij}^{\,n}.
  \label{eq:method-corotated-deformation-measure}
\end{equation}
Here \(\bm{I}\) is the identity tensor and
\(\bm{F}_{i}^{\mathrm{co},n}\) denotes the deformation measure expressed in the
local corotational frame.  Thus the same first-order corrected SPH gradient
provides both the local deformation measure and the rotation-aware force
discretization.  At each time step, stresses are evaluated at all supports, and
the internal forces are assembled by the corrected stress-divergence operator.
The same two-pass stress and force construction is also used to evaluate the
matrix-free action required by the implicit solver.

The constitutive law specifies the first Piola--Kirchhoff stress associated
with the corrected deformation state at each support,
\(\bm{P}_i=\widehat{\bm{P}}(\bm{F}_i,\dot{\bm{F}}_i;
\bm{\theta}_m)\), where \(\dot{\bm{F}}_i\) denotes the material
time rate of the deformation gradient and \(\bm{\theta}_m\)
collects the material parameters.  The same corrected deformation-gradient
operator and corrected stress-divergence operator are therefore used for all
constitutive choices considered below.  Young's modulus \(E\) and Poisson's
ratio \(\nu\) define the Lam\'e parameters
\(\mu=E/[2(1+\nu)]\) and
\(\lambda=E\nu/[(1+\nu)(1-2\nu)]\), which are used by the elastic responses.
The operators
\(\operatorname{sym}(\bm{A})=(\bm{A}+\bm{A}^T)/2\),
\(\operatorname{tr}(\bm{A})\), and
\(\operatorname{dev}(\bm{A})=\bm{A}-\operatorname{tr}(\bm{A})\bm{I}/3\)
denote the symmetric, trace, and deviatoric parts of a second-order tensor.

For the corotated linear material, denoted by the superscript
\(\mathrm{CL}\), the strain is evaluated in the corotational frame and then
rotated back to the reference stress measure:
\begin{equation}
  \bm{\varepsilon}_i
  =
  \operatorname{sym}(\bm{R}_i^T\bm{F}_i)-\bm{I},
  \qquad
  \bm{P}_i^{\mathrm{CL}}
  =
  \bm{R}_i
  \left[
    2\mu\bm{\varepsilon}_i
    +
    \lambda\operatorname{tr}(\bm{\varepsilon}_i)\bm{I}
  \right].
  \label{eq:method-corotated-linear}
\end{equation}

For compressible logarithmic Neo--Hookean elasticity, denoted by the
superscript \(\mathrm{NH}\), the first Piola--Kirchhoff stress is evaluated
directly from \(\bm{F}_i\):
\begin{equation}
  \bm{P}_i^{\mathrm{NH}}
  =
  \mu\bm{F}_i
  +
  (\lambda\ln J_i-\mu)\bm{F}_i^{-T},
  \qquad
  J_i=\det\bm{F}_i>0.
  \label{eq:method-neo-hookean}
\end{equation}
Unlike the corotated material, this finite-strain model does not introduce a
separate rotation extraction in the stress evaluation.

Kelvin--Voigt viscosity augments either elastic response with a
rate-dependent viscous stress \(\bm{P}_i^v\).  For the finite-strain
formulation,
\begin{equation}
  \dot{\bm{E}}_i
  =
  \operatorname{sym}(\bm{F}_i^T\dot{\bm{F}}_i),
  \qquad
  \bm{P}_i^{v}
  =
  \bm{F}_i
  \left[
    2\eta_s\operatorname{dev}(\dot{\bm{E}}_i)
    +
    \eta_b\operatorname{tr}(\dot{\bm{E}}_i)\bm{I}
  \right].
  \label{eq:method-kv-stress}
\end{equation}
The corotated material uses the analogous strain rate in its local frame.  The
total stress is then assembled as
\(\bm{P}_i=\bm{P}_i^e+\bm{P}_i^v\), where \(\bm{P}_i^e\) denotes the elastic
part, either
\(\bm{P}_i^{\mathrm{CL}}\) or \(\bm{P}_i^{\mathrm{NH}}\).  Both elastic and
viscous stresses enter the same corrected divergence operator, which allows the
rate-dependent stress to be added without changing the Lagrangian SPH
kinematics.  The material parameters \(\eta_s\) and \(\eta_b\) are distinct
from the numerical velocity damping used only to relax quasi-static
simulations.

\subsection{Matrix-free implicit solution}
\label{sec:method-time-integration}

The discrete mechanical residual \(\mathcal Q(\bm z,\bm\theta)\) in
Eq.~\eqref{eq:method-forward-residual} comprises the momentum equations and
position updates over the simulated trajectory.  The boundary forces and matrix-free
implicit solution defining this residual are detailed below.

Dirichlet conditions prescribe the support displacement
\(\bm u_i=\bm x_i-\bm X_i=\overline{\bm u}_i\), and Neumann conditions are
represented by applied nodal forces.  At each support \(i\) on the Robin boundary,
partial restraint is represented by the restoring force
\begin{equation}
  \bm{f}_{i}^{R}
  =
  -m_i k_i(\bm{u}_i-\overline{\bm{u}}_i),
  \label{eq:method-robin-force}
\end{equation}
where \(k_i\geq0\) is the support stiffness and
\(\overline{\bm u}_i\) is its target displacement.  Zero stiffness gives
no restraint, while increasing stiffness approaches the prescribed
displacement.

The corrected solid-SPH model is advanced by an implicit velocity update.  At
the beginning of time step \(n\), the deformation gradients, local rotations,
and rotated corrected gradients are evaluated from the current configuration.
They are then held fixed during the velocity solve, following the linearized
implicit treatment of corrected corotated SPH.  With the position update
\(\bm{x}^{n+1}=\bm{x}^{n}+\Delta t\,\bm{v}^{n+1}\), the semi-implicit
velocity residual can be written as
\begin{equation}
  \bm{v}^{n+1}
  -
  \bm{v}^{n}
  -
  \Delta t\,\bm{M}^{-1}
  \bm{f}^{\,n}(\bm{v}^{n+1};\bm{\theta})
  =
  \bm{0},
  \label{eq:method-implicit-residual}
\end{equation}
where \(\bm{M}\) is the mass matrix and \(\bm{f}^{\,n}\) denotes the force
model evaluated with the
step-frozen kinematic quantities.  Linearization with respect to velocity
gives the implicit velocity system and position update
\begin{equation}
  \mathcal{A}^{n}(\bm{\theta})\bm{v}^{n+1}
  =
  \bm{b}^{n}(\bm{\theta}),
  \qquad
  \bm{x}^{n+1}
  =
  \bm{x}^{n}
  +\Delta t\,\bm{v}^{n+1},
  \label{eq:method-linear-system}
\end{equation}
where \(n\) is the time-step index, \(\Delta t\) is the time-step size, and
\(\bm x^n\) and \(\bm v^n\) collect the support positions and velocities.
The operator \(\mathcal A^n\) and right hand side \(\bm b^n\) depend on the
current state and physical parameters \(\bm\theta\).
The velocity system's left hand side includes the
linearized elastic, viscous, and Robin forces, while its right hand side
contains the known velocity and current forces.  The operator is evaluated
by its action on a trial velocity field \(\bm p\), using conjugate gradients
without assembling the matrix.  Numerical velocity damping may be applied
for quasi-static relaxation and is distinct from the material viscosity.

Let \(\bm{f}_{e}^{\,n}\) denote the elastic internal force evaluated at
\(\bm{x}^{n}\), \(\delta\bm{f}_{e}^{\,n}[\Delta t\bm{p}]\) the corresponding
elastic force increment induced by the trial displacement \(\Delta t\bm{p}\),
\(\bm{f}_{v}^{\,n}[\bm{p}]\) the Kelvin--Voigt viscous force induced by the
trial velocity, and
\(\delta\bm{f}_{R}^{\,n}[\Delta t\bm{p}]\) the Robin-like support-force
increment.  The left hand side operator in
Eq.~\eqref{eq:method-linear-system} is defined through its action on an
arbitrary trial velocity field \(\bm{p}\):
\begin{equation}
  \left(\mathcal{A}^{n}\bm{p}\right)_i
  =
  \bm{p}_i
  -
  \Delta t\,m_i^{-1}
  \left(
    \delta\bm{f}_{e,i}^{\,n}[\Delta t\bm{p}]
    +
    \bm{f}_{v,i}^{\,n}[\bm{p}]
    +
    \delta\bm{f}_{R,i}^{\,n}[\Delta t\bm{p}]
  \right).
  \label{eq:method-matrix-free-operator}
\end{equation}
The scaling in Eq.~\eqref{eq:method-matrix-free-operator} follows from the
physical variables on which each force depends.  The elastic and Robin-like
contributions are displacement-dependent and therefore act on the trial
displacement \(\Delta t\bm{p}\).  The Kelvin--Voigt contribution is
rate-dependent and acts on the trial velocity \(\bm{p}\).  Differentiating
the Robin force in Eq.~\eqref{eq:method-robin-force} with respect to
displacement, with stiffness and target displacement held fixed, gives
the increment at each Robin support \(i\):
\begin{equation}
  \delta\bm{f}_{R,i}^{\,n}[\Delta t\bm{p}]
  =
  -m_i k_i\Delta t\,\bm{p}_i.
  \label{eq:method-robin-matvec}
\end{equation}
Thus a positive Robin stiffness contributes a positive resistance in the
linearized implicit operator after substitution into
Eq.~\eqref{eq:method-matrix-free-operator}.

The right hand side of Eq.~\eqref{eq:method-linear-system} contains the known
velocity and the current forces,
\begin{equation}
  \bm{b}_i^{n}
  =
  \bm{v}_i^n
  +
  \Delta t\,m_i^{-1}
  \left(
    \bm{f}_{e,i}^{\,n}
    +
    \bm{f}_{R,i}^{\,n}
    +
    \bm{f}_{L,i}^{\,n}
  \right),
  \label{eq:method-implicit-rhs}
\end{equation}
where \(\bm{f}_{R,i}^{\,n}\) is given by Eq.~\eqref{eq:method-robin-force} and
\(\bm{f}_{L,i}^{\,n}\) denotes the prescribed Neumann load.  In the
semi-implicit Kelvin--Voigt treatment, the rate-dependent viscous response is
included through the operator action in
Eq.~\eqref{eq:method-matrix-free-operator}.

Following the two-pass corrected SPH force evaluation of
Peer et al.~\cite{peer2018implicit}, each current force and each matrix-vector
product are evaluated by the same traversal of the fixed reference neighbor
graph.  The first pass
constructs the deformation-gradient, strain-rate, stress, or stress-increment
quantity at every support, depending on the selected constitutive model.  The
second pass applies the corrected stress-divergence operator to assemble the
corresponding force or force increment.  Consequently, neo-Hookean elasticity modifies the stress and material tangent while
preserving the corrected SPH force assembly, Kelvin--Voigt viscosity contributes
a rate-dependent stress through \(\dot{\bm{F}}\), and the Robin-like condition
contributes both a current restoring force and a tangent resistance to the same
implicit solve.

\FloatBarrier


\section{Numerical verification of the 3DGS–SPH model components}
\label{app:numerical-verification}
\setcounter{figure}{0}
\setcounter{table}{0}

The following studies support the forward simulation and inverse analysis
in Section~4 by examining the corrected SPH response, the Lagrangian
SPH--GS transfer, and the objective gradients used for optimization.
The benchmark configurations, comparison procedures, and detailed results
are collected here.

\subsection{Validation of the solid SPH discretization}
\label{sec:exp01-sph-fem-reference}

This subsection provides a reference check for the solid SPH discretization. 
A linear elastic beam is used
as a controlled benchmark, and the adopted corrected SPH response is compared
with a finite-element reference computed using JAX-FEM~\cite{xue2023jaxfem}.
The comparison examines whether the SPH discretization reproduces the dominant
deformation pattern of the reference solution.

The beam dimensions are \(L_x=2.0\,\mathrm{m}\) and
\(L_y=L_z=0.5\,\mathrm{m}\). Both models use \(E=10^4\,\mathrm{Pa}\) and
\(\nu=0.35\) for quasi-static forward simulation. The left end and the bottom
segment \(x/L_x\in[0,0.2]\) are constrained, while a downward resultant
\(\|\bm{f}\|=20\,\mathrm{N}\) is applied at the right end. No
Robin-type support is active in this reference case. The FEM reference employs
\(40\times10\times10\) hexahedral elements on the same represented domain, with
a uniform right-face pressure whose integral equals the SPH resultant load.

Four models are included in the comparison. The first is the JAX-FEM reference.
The second is the adopted corrected SPH model, represented by a
\(40\times10\times10\) cell-centered lattice with spacing
\(\Delta=0.05\,\mathrm{m}\) and smoothing length \(h=0.075\,\mathrm{m}\).
Assigning each of the 4000 particles the finite volume \(\Delta^3\) gives
\(\sum_iV_i^0=L_xL_yL_z=0.5\,\mathrm{m^3}\). The third model is a corrected surface-dense SPH model, where 1000 additional supports are
added on the six beam faces. The fourth model uses the same particle positions, masses, and
boundary data as the adopted SPH model, but replaces the first-order corrected
kernel gradient \(\widehat{\nabla}_iW_{ij}^0=\bm{L}_i\nabla_iW_{ij}^0\) by the
uncorrected gradient \(\nabla_iW_{ij}^0\), where \(\bm{L}_i\) is the local
correction matrix defined in \ref{sec:method-corrected-sph},
Eq.~\eqref{eq:method-gradient-correction}.

For the surface-dense model, the total represented material is kept
unchanged. Let \(\mathcal S\) denote the original surface supports,
\(\mathcal S^a\) the added supports, and
\(\mathcal S'=\mathcal S\cup\mathcal S^a\) the dense surface set. If
\(M_{\mathcal S}=\sum_{i\in\mathcal S}m_i\) is the original surface mass, the
interior-support masses are retained and each member of \(\mathcal S'\) is
assigned
\begin{equation}
  m_i'=\frac{M_{\mathcal S}}{|\mathcal S'|},\qquad
  V_i^{0\prime}=\frac{m_i'}{\rho_0},\qquad i\in\mathcal S' .
  \label{eq:surface-support-mass-redistribution}
\end{equation}
This redistribution changes the surface-support distribution without changing
the total mass or rest volume.

\begin{figure}[!htbp]
  \centering
  \includegraphics[width=0.6\linewidth]{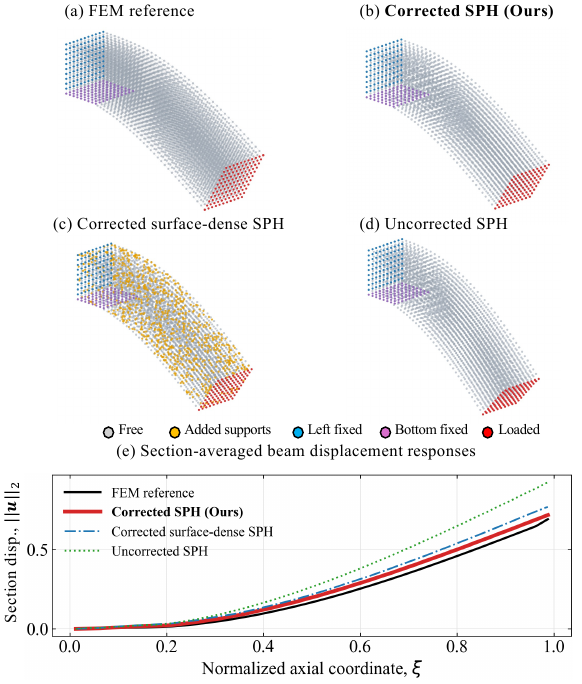}
  \caption{Reference validation of the solid SPH discretization. Panels
  (a)--(d) show the final deformed configurations of the FEM reference, the
  corrected SPH model (ours), the corrected surface-dense SPH model, and
  the uncorrected SPH model, respectively. Panel (e) compares the
  section-averaged displacement magnitude along the normalized axial
  coordinate \(\xi=(X_x-X_{\min})/(X_{\max}-X_{\min})\).}
  \label{fig:exp01-sph-fem-validation}
\end{figure}

The final deformed configurations and the section-averaged displacement
responses are summarized in Fig.~\ref{fig:exp01-sph-fem-validation}. The adopted
corrected SPH model follows the global bending pattern of the FEM reference,
whereas the surface-dense and uncorrected variants produce visibly larger
deflections. This comparison indicates that both the spatial support
distribution and the first-order kernel correction affect the beam response.

\FloatBarrier

Because the SPH particles and FEM nodes do not coincide, the response is
compared through axial sectional averages of the displacement magnitude \(\|\bm{u}\|_2\), where
\(\bm{u}\) is the displacement vector. The axial profiles in
Fig.~\ref{fig:exp01-sph-fem-validation}(e) show that the adopted corrected SPH
model remains closest to the FEM reference among the three SPH models.

Table~\ref{tab:exp01-sph-fem-reference} summarizes the same comparison using
40 axial sections. The normalized root mean square error (NRMSE) is computed
relative to the maximum FEM displacement-magnitude response. For the adopted
corrected SPH model, the displacement-magnitude NRMSE relative to FEM is
\(4.27\%\), and the maximum displacement magnitude is close to the FEM
reference. The surface-enriched and uncorrected models show larger errors,
supporting the use of the corrected, regularly distributed SPH support in the
subsequent 3DGS-SPH coupling experiments.

\begin{table}[!htbp]
  \centering
  \small
  \caption{Comparison of the FEM reference and SPH models for the linear beam
  benchmark.}
  \label{tab:exp01-sph-fem-reference}
  \begin{tabular}{l c c c}
    \toprule
    Model & Supports/nodes & Maximum \(\|\bm{u}\|_2\) (m) &
    \(\|\bm{u}\|_2\) NRMSE \\
    \midrule
    FEM reference & 4961 nodes & 0.7183 & -- \\
    Corrected SPH & 4000 supports & 0.7343 & \(4.27\%\) \\
    Surface-enriched corrected SPH & 5000 supports & 0.7900 & \(8.16\%\) \\
    Uncorrected SPH & 4000 supports & 0.9543 & \(18.9\%\) \\
    \bottomrule
  \end{tabular}
\end{table}
\FloatBarrier

Overall, this reference comparison supports the corrected SPH discretization
used in the remaining experiments. It also indicates that surface-biased
support distributions and uncorrected kernel gradients are less suitable as the
mechanical basis of the coupled 3DGS-SPH model.

\FloatBarrier

\subsection{Validation of the Lagrangian SPH-GS transfer}
\label{sec:exp00b-lagrangian-map}

This experiment evaluates the agreement between the deformation renderings
produced by the proposed Lagrangian-coupled 3DGS-SPH model and the conventional co-located model.  The benchmark 3DGS rendering model is reconstructed from the beam with DIC-style speckle texture mentioned in Section~\ref{sec:exp10-forward-beam10k}
(Fig.~\ref{fig:exp00b-reconstruction}).

\begin{figure}[!htbp]
  \centering
  \includegraphics[width=0.98\linewidth]{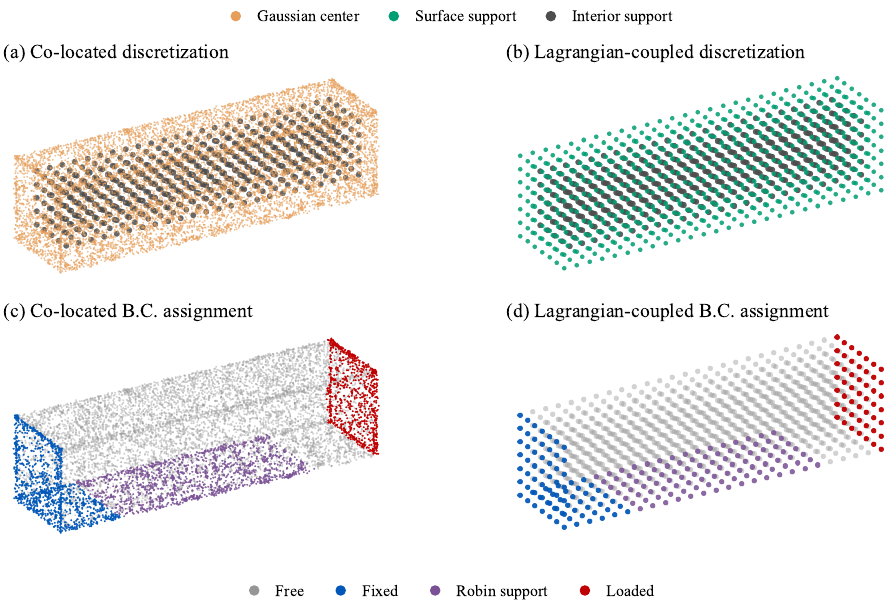}
  \caption{Physical discretizations and B.C. assignments used in the transfer
  experiment.  (a) Co-located model, in which every surface support coincides
  with a Gaussian center and is therefore not overplotted.  (b) Proposed
  Lagrangian-coupled model with an independently discretized SPH domain;
  Gaussian centers are omitted to expose its support distribution.  Panels
  (c) and (d) show the fixed, weak Robin-supported, loaded, and free supports
  selected in the two models.}
  \label{fig:exp00b-configuration}
\end{figure}

Two coupled constructions are considered.  The conventional co-located model
assigns one surface SPH support to every Gaussian center, giving 10,000 surface
supports supplemented by 600 interior supports.  This direct correspondence
between graphical and physical primitives is commonly adopted in
physics-integrated Gaussian representations
\cite{xie2024physgaussian,xiong2025topogaussian}.  In contrast, the proposed
Lagrangian-coupled model constructs the physical discretization independently,
using 674 quasi-uniform surface supports and the same 600 interior supports.
The resulting 1,274-particle state drives all 10,000 Gaussian kernels through
the reference-configuration transfer graph introduced in
Sec.~\ref{sec:method-lagrangian-transfer}. These two physical discretizations
are compared in Fig.~\ref{fig:exp00b-configuration}(a) and (b).

The co-located physical discretization inherits the surface-concentrated Gaussian kernels of 3DGS model, which is not designed
to provide the volumetric neighborhoods required by SPH quadrature.  It also
causes the number of physical degrees of freedom to grow directly with the
dimension of the 3DGS model.  Because the two constructions contain different numbers
of surface supports, the redistribution defined in
Eq.~\eqref{eq:surface-support-mass-redistribution} is applied to the co-located
model while retaining its interior discretization, so that the two models consequently
have the same represented mass and rest volume.

Both models are advanced from rest for 1,000 steps the with the corrected SPH formulation. The corotated linear elastic material parameters are
\(E=10^4\,\mathrm{Pa}\), \(\nu=0.35\), and
\(\rho_0=1200\,\mathrm{kg}\cdot\mathrm{m^{-3}}\).  A smoothing length of
\(h=0.10\,\mathrm{m}\), time increment \(\Delta t=0.02\,\mathrm{s}\), and
dimensionless per-step velocity-damping factor of 0.02 are used to obtain a
quasi-static deformation response.  The left end
and the lower surface over the first 20\% of
the beam are fixed; the remaining designated lower patch has weak Robin
support with mass-normalized stiffness
\(k_R=10^{-2}\,\mathrm{s^{-2}}\), and a downward resultant load of
magnitude \(20\,\mathrm{N}\) is
applied at the right end.  Figure~\ref{fig:exp00b-configuration} (c) and (d) show the
particle masks produced by this common boundary condition prescription.
The co-located construction and the proposed Lagrangian SPH--GS transfer are
compared using the same 3DGS rendering model.  Their underlying SPH
discretizations, however, would produce different displacement amplitudes as mentioned in \ref{sec:exp01-sph-fem-reference} (surface-dense SPH model case).  To
reduce the influence of this difference on the rendering comparison, the
Gaussian-center displacements and affine-deformation increments obtained
through the Lagrangian transfer are scaled to match the mean loaded-end
Gaussian displacement of the co-located construction.  The rendered images
are then compared at matched deformation amplitudes.

Agreement between the two mapping methods is quantified by global and
pointwise errors in the Gaussian-center displacements of the 3DGS model.  Let
\(\bm u_g^{m}(t)=\bm\mu_g^{m}(t)-\bm\mu_g^0\) denote the displacement of
Gaussian \(g\) from its rest center \(\bm\mu_g^0\), where the superscripts
\(\mathrm{col}\), \(\mathrm{Lag}\), and \(\mathrm{adj}\) identify the
co-located response and the Lagrangian-transfer response before and after amplitude matching,
respectively, and \(N_G\) is the total number of Gaussian kernels.  Taking the
co-located Gaussian displacements as the comparison baseline, these error
measures are defined as
\[
  \varepsilon_2^{m}(t)
  =
  \frac{
    \left[\sum_{g=1}^{N_G}
    \|\bm u_g^{m}(t)-\bm u_g^{\mathrm{col}}(t)\|_2^2\right]^{1/2}}
  {
    \left[\sum_{g=1}^{N_G}\|\bm u_g^{\mathrm{col}}(t)\|_2^2\right]^{1/2}},
  \quad
  e_g^{m}(t)
  =
  \frac{
    \|\bm u_g^{m}(t)-\bm u_g^{\mathrm{col}}(t)\|_2}
  {
    \left[N_G^{-1}\sum_{q=1}^{N_G}
    \|\bm u_q^{\mathrm{col}}(t)\|_2^2\right]^{1/2}},
  \quad m\in\{\mathrm{Lag},\mathrm{adj}\}.
\]
Here \(\varepsilon_2^m\) measures the global relative error, whereas
\(e_g^m\) describes its spatial distribution over individual Gaussians.

\begin{figure}[!htbp]
  \centering
  \includegraphics[width=0.99\linewidth]{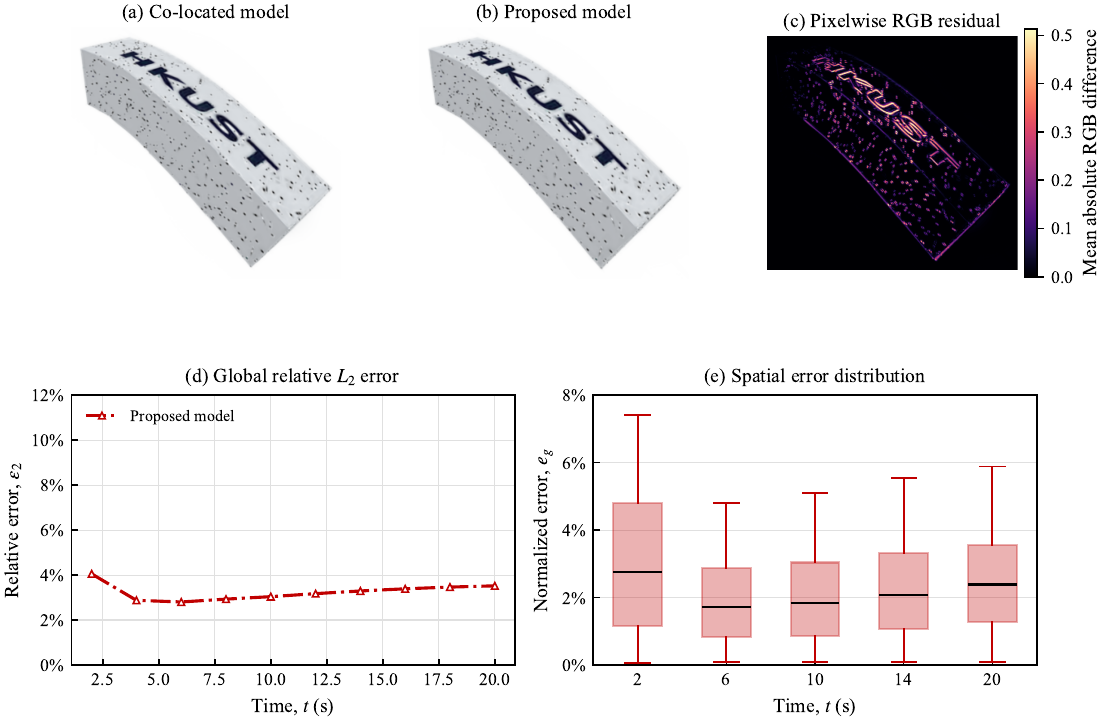}
  \caption{Rendering and Gaussian-displacement agreement between the
  co-located construction and the proposed 3DGS--SPH model with Lagrangian
  SPH--GS transfer.  The comparison uses matched deformation amplitudes.
  (a) Co-located model.  (b) Proposed model.
  (c) Pixelwise mean absolute RGB difference between (a) and (b), with the
  color scale truncated at the 99.5th percentile.
  (d) Global relative \(L_2\)
  Gaussian-displacement error \(\varepsilon_2(t)\) of the proposed model over simulation time.
  (e) Pointwise normalized displacement errors \(e_g\) at five representative
  states.  Both displacement-error measures use the co-located response as
  the baseline; boxes denote the interquartile range, center lines the median,
  and whiskers the 5th and 95th percentiles.}
  \label{fig:exp00b-transfer-results}
\end{figure}

Figure~\ref{fig:exp00b-transfer-results}(a) and (b) compare the final
renderings obtained from the co-located construction and the proposed
3DGS--SPH model, for which
the image-space RMSE is \(3.784\times10^{-2}\).  The residual field in
Fig.~\ref{fig:exp00b-transfer-results}(c) shows that the remaining differences
are concentrated near the projected boundary and textured features.
Figure~\ref{fig:exp00b-transfer-results}(d) tracks the global relative
\(L_2\) error in the proposed model's Gaussian-center displacements with
respect to the co-located response over simulation time, assessing their
agreement along the deformation trajectory after amplitude matching.
The pointwise distributions in
Fig.~\ref{fig:exp00b-transfer-results}(e) further show that at least 75\% of
the Gaussian kernels have \(e_g<5\%\), while the 95th percentile remains below
7.5\%.

The proposed model reduces the physical particle count by
88.0\% and the number of SPH neighbor pairs from 3,394,870 to 72,370.  On the
same hardware, the 1,000-step rollout time decreases from
\(497.7\,\mathrm{s}\) to \(10.37\,\mathrm{s}\), corresponding to a 48.0-fold
acceleration.  Together with the bounded transfer error and close agreement
between the amplitude-adjusted deformation renderings, these results
demonstrate that the proposed Lagrangian coupling can substantially reduce the
physical simulation cost while retaining the deformation characteristics
required for Gaussian rendering.

\FloatBarrier

\subsection{Automatic differentiation validation of the 3DGS-SPH coupled pipeline}
\label{sec:exp20-ad-validation}

This experiment evaluates the accuracy of sensitivities computed by automatic
differentiation (AD) using the beam model in Section~\ref{sec:exp10-forward-beam10k}.  The evaluated objective is the image-mismatch term \(\mathcal J_{\mathrm{o}}\) in
Eq.~\eqref{eq:method-image-objective}.  The reference images are generated using known
physical parameters.  Differentiation includes the SPH simulation,
Lagrangian SPH--GS transfer, Gaussian covariance update, and rendering.

Directional derivatives obtained by AD are compared with central finite-difference (FD) estimates along the same unit parameter direction.  Relative error is the absolute difference between the AD and FD estimates divided by the larger of their absolute magnitudes.
The SPH neighbor graph, SPH--GS transfer graph, boundary masks, and Gaussian ordering remain fixed during each comparison.

\begin{figure}[!htbp]
  \centering
  \includegraphics[width=0.78\linewidth]{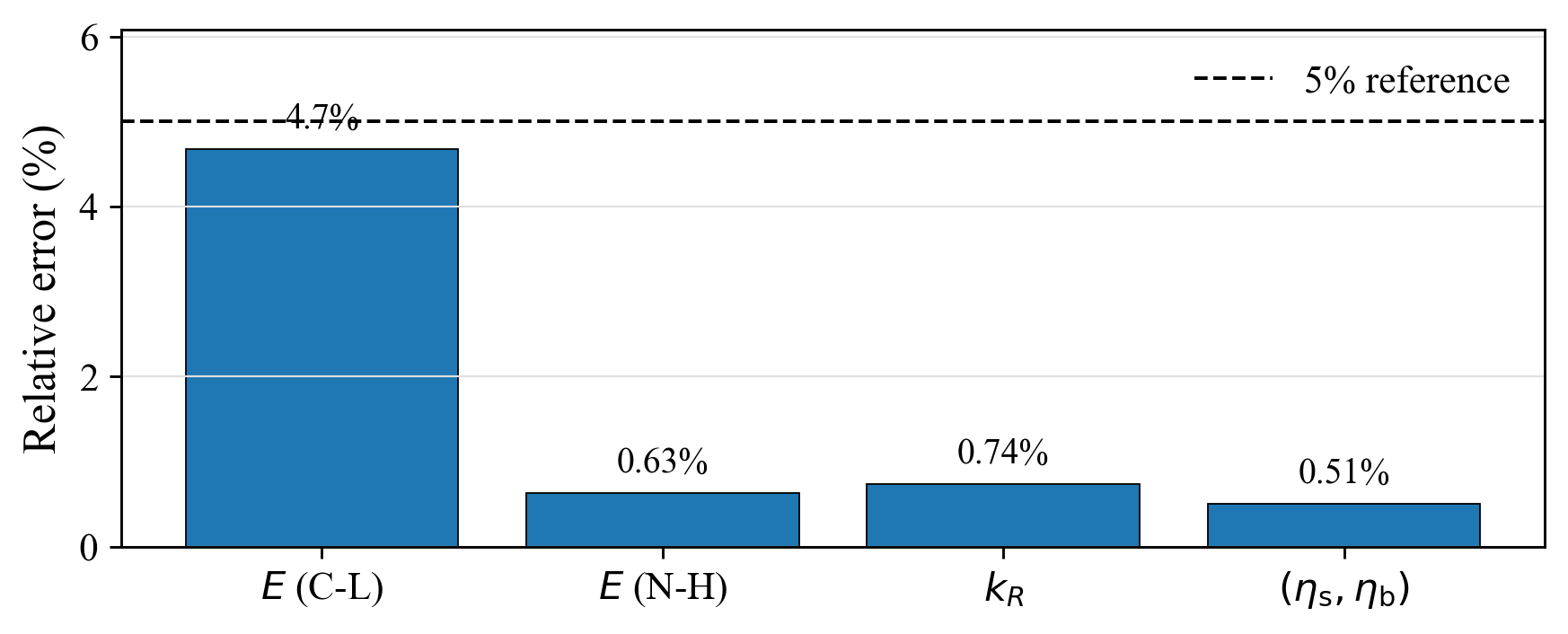}
  \caption{Automatic differentiation validation for the beam 3DGS--SPH coupled
  pipeline.  Relative errors between AD and FD directional derivatives of
  \(\mathcal J_{\mathrm{o}}\) are evaluated through the
  complete SPH rollout, SPH--GS transfer, covariance update, and
  differentiable rendering path.  Here \(E_{\mathrm{C\text{-}L}}\) and
  \(E_{\mathrm{N\text{-}H}}\) denote Young's-modulus checks for the
  corotated-linear and neo-Hookean material models, respectively.}
  \label{fig:exp20-ad-dashboard}
\end{figure}

The tested parameters comprise Young's modulus for corotated-linear and
neo-Hookean elasticity, the Robin stiffness field, and Kelvin--Voigt
viscosity.  Table~\ref{tab:exp20-ad-inversestart} lists the evaluation values,
at which derivatives are computed, and the reference values used to generate
the images.  These values differ so that sensitivities are assessed away
from the reference solution.  For vector-valued parameters, each listed
value is assigned to all components.  Static material and boundary tests
use one rendered terminal frame, whereas the viscosity test uses the last
five rendered frames of a transient simulation.  The FD perturbation
magnitude is \(10^{-3}\), except for Young's modulus, for which \(10^{-2}\)
is used.  The resulting relative errors are reported in
Table~\ref{tab:exp20-ad-inversestart} and
Fig.~\ref{fig:exp20-ad-dashboard}.

\begin{table}[!htbp]
  \centering
  \small
  \caption{AD--FD comparison for the image-mismatch term
  \(\mathcal J_{\mathrm{o}}\) in Eq.~\eqref{eq:method-image-objective}.
  Evaluation values specify where derivatives are computed; reference values
  generate the reference images.  Dim. denotes the parameter dimension.}
  \label{tab:exp20-ad-inversestart}
  \setlength{\tabcolsep}{4pt}
  \begin{tabular}{l l r r r r}
    \toprule
    Parameter (unit) &
    Material/B.C. field &
    Evaluation &
    Reference &
    Dim. &
    Relative error \\
    \midrule
    \(E\) (Pa) & corotated linear & 5000 & 10000 &
    1 &
    \(4.6788\times10^{-2}\) \\
    \(E\) (Pa) & neo-Hookean & 5000 & 10000 &
    1 &
    \(6.3272\times10^{-3}\) \\
    \(k_R\) (\(\mathrm{s^{-2}}\)) & Robin stiffness field & 1000 & 100 &
    105 &
    \(7.4073\times10^{-3}\) \\
    \((\eta_s,\eta_b)\) (\(\mathrm{Pa}\cdot\mathrm{s}\)) & Kelvin--Voigt & 1000 & 5000 &
    2 &
    \(5.0778\times10^{-3}\) \\
    \bottomrule
  \end{tabular}
\end{table}
\FloatBarrier

\FloatBarrier

All selected checks remain below the \(5\%\) relative-error reference.  The
largest discrepancy occurs for the corotated-linear Young's-modulus check,
where the rotation extracted from the deformation gradient is recomputed in
the forward simulation but treated as a frozen auxiliary quantity in the
backward sensitivity calculation for numerical stability.  This controlled
approximation does not enter the neo-Hookean material derivative, which is
evaluated directly from \(\bm{F}\), and the corresponding sensitivity error is
therefore smaller.  The Robin and Kelvin--Voigt rows show close agreement
between automatic differentiation and finite differences, supporting the use
of the rendered MSE objective for the inverse analyses in
Section~\ref{sec:exp30-inverse-recovery}.

\FloatBarrier

\section{Notation}
\label{app:notation}

Table~\ref{tab:method-notation} collects notation used across multiple method
subsections.  Scalars are italic, vectors are bold lowercase, and matrices and
second-order tensors are bold uppercase.  Superscript $0$ denotes the
reference configuration, while superscript $n$ denotes simulation step
$t_n=n\Delta t$.  Symbols used only locally are defined with their equations.

\begin{table}[H]
  \centering
  \small
  \caption{Principal notation used in the coupled formulation.}
  \label{tab:method-notation}
  \begin{tabular}{@{}p{0.22\linewidth}p{0.7\linewidth}@{}}
    \toprule
    Symbol & Definition \\
    \midrule
    $g=1,\ldots,G$ & Gaussian index and number of rendered Gaussians. \\
    $i,j=1,\ldots,N$ & SPH support indices and number of physical supports. \\
    \(\bm{\mu}_g^0,\bm{\mu}_g^n\) & Reference and current Gaussian centers. \\
    \(\bm{\Sigma}_g^0,\bm{\Sigma}_g^n\) & Reference and current Gaussian covariances. \\
    \(\bm{X}_i,\bm{x}_i^n,\bm{v}_i^n\) & Reference position, current position, and velocity of SPH support $i$. \\
    \(m_i,\Delta V_i,\rho_0\) & Support mass, reference volume, and reference density. \\
    \(\bm{F}_i^n,\bm{F}_g^n\) & SPH- and Gaussian-level deformation gradients. \\
    $h,h_{\mathrm{map}}$ & Mechanical smoothing length and transfer smoothing length. \\
    \(\mathcal{E}_{\mathrm{sph}}\) & Fixed reference graph used by the SPH mechanics. \\
    \(\mathcal{E}_{\mathrm{map}}\) & Fixed bipartite graph from Gaussians to supporting SPH particles. \\
    $E,\nu,\eta_s,\eta_b$ & Young's modulus, Poisson ratio, and shear and bulk viscosity parameters. \\
    \(\bm{\theta}\) & Material and boundary parameters selected for inverse identification. \\
    \bottomrule
  \end{tabular}
\end{table}

\bibliographystyle{unsrt}
\bibliography{references}

\end{document}